\documentclass{article}

\PassOptionsToPackage{numbers}{natbib}
\usepackage[main, final]{neurips_2026}

\newcommand\clgray[1]{{\color{gray}{#1}}}
\newcommand{\settablefont}{\fontsize{6}{9}\selectfont}

\newcommand{\hatsup}[2]{{\smash[t]{\hat{#1}}\vphantom{#1}}^{#2}%
}

\usepackage{caption}  %
\usepackage{xcolor}   %

\usepackage{graphicx}  %
\usepackage{pifont}

\usepackage{arydshln}

\usepackage[utf8]{inputenc} %
\usepackage[T1]{fontenc}    %
\usepackage{hyperref}       %
\usepackage{url}            %
\usepackage{booktabs}       %
\usepackage{amsfonts}       %
\usepackage{microtype}      %

\usepackage{amsmath,amssymb} %
\usepackage{multirow}

\usepackage{marvosym}

\newcommand{\authorskip}{\hspace{3mm}}
\renewcommand{\thefootnote}{\Letter}

\hypersetup{
    colorlinks=true,
    linkcolor=black,
    citecolor=green!50!black,
    urlcolor=magenta
}

\title{Learning to Discriminate Scene Structures Makes Self-Supervised Depth Learning Scalable}

\usepackage{etoolbox}

\makeatletter
\patchcmd{\@maketitle}
  {\rule{\z@}{24\p@}\@author}
  {\rule{\z@}{12\p@}\@author}
  {}
  {\PackageWarning{author-spacing}{Patch failed}}
\makeatother

\author{
Mengtan Zhang\textsuperscript{1} \authorskip
Yuchuan Cui\textsuperscript{2} \authorskip
Yu Ma\textsuperscript{2} \authorskip
Zizhan Guo\textsuperscript{2} \authorskip
Wei Ye\textsuperscript{1,2} \authorskip
Rui Fan\textsuperscript{1,2,3,\Letter} \\[4pt]
\parbox{\dimexpr\textwidth-2\tabcolsep\relax}{%
    \centering\normalfont\small
    \textsuperscript{1}Shanghai Research Institute for Intelligent
    Autonomous Systems, Tongji University\\
    \textsuperscript{2}College of Electronic and Information Engineering,
    Tongji University\\
    \textsuperscript{3}State Key Laboratory of Human-Machine
    Hybrid Augmented Intelligence, Xi'an Jiaotong University\\[3pt]
    \texttt{zmt200110@tongji.edu.cn},
    \texttt{rui.fan@ieee.org}
}
}

\begin{document}

\maketitle

\renewcommand{\thefootnote}{\arabic{footnote}}

\let\thefootnote\relax
\footnotetext{
\small
\textsuperscript{\Letter} Corresponding author.
}

\begin{center}
\vspace{-2.5em}
    \includegraphics[width=0.975\textwidth]{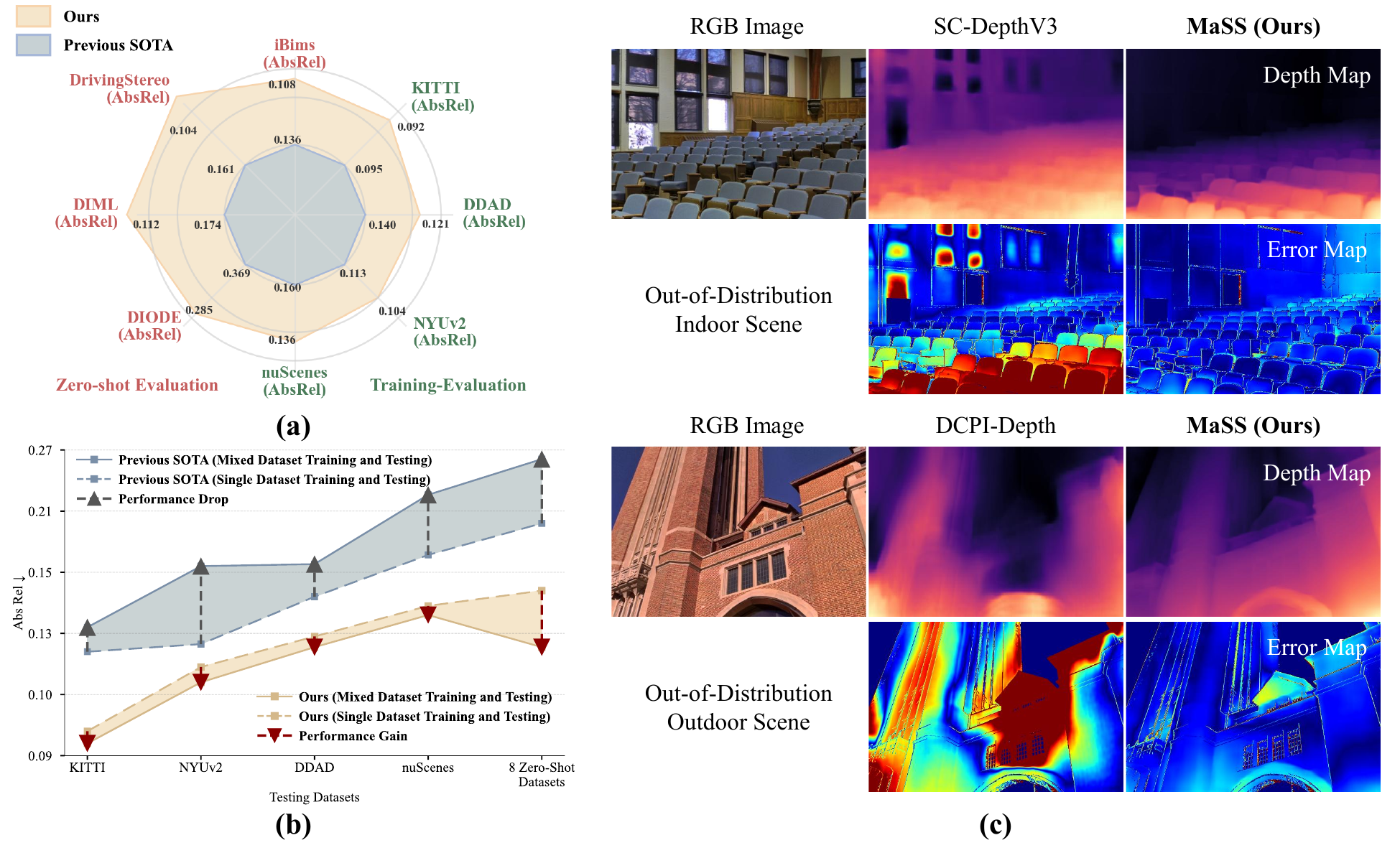}
    \captionof{figure}{
    Beyond (a) achieving SoTA performance across diverse datasets with a unified model, MaSS is the first SSMDE method to (b) effectively benefit from increased data diversity and (c) exhibit remarkable generalizability to entirely out-of-distribution scenes.}
    \label{fig.teaser}
\end{center}

\begin{abstract}
  Self-supervised monocular depth estimation is appealing for its potential to scale with unlabeled data, yet existing methods often degrade when trained on diverse scenes. This study finds that, under self-supervision, models tend to encode dataset-specific structural biases rather than transferable scene geometry, leading to cross-domain optimization conflicts. To address this issue, this study proposes the MaSS framework, aiming to make self-supervised depth learning scalable to diverse data. MaSS introduces a dedicated structural representation pathway explicitly decoupled from contextual features, which is guided by a novel scene-structure discrimination objective to learn generalized, layout-aware structural representations for depth decoding. Trained and evaluated on four diverse datasets, both individually and jointly, MaSS substantially outperforms prior self-supervised methods and is the first to consistently benefit from mixed-domain training. Zero-shot evaluation on eight additional datasets further demonstrates its remarkable generalizability, particularly on those entirely out-of-distribution scenes.
  The source code is publicly available at \url{https://mias.group/MaSS}.
\end{abstract}

\section{Introduction}

The fundamental advantage of self-supervised learning lies in its potential to leverage ubiquitous unlabeled data for scalable model training. However, most existing self-supervised monocular depth estimation (SSMDE) research \cite{godard2019digging, zhao2023gasmono, sun2023sc, zhang2024dcpi, zhang2025discriminately} is predominantly confined to training models individually on single datasets or scenarios, leaving much of this potential unexplored. Recent efforts have attempted to directly train models with existing SSMDE methods on large-scale video sequences curated from online platforms \cite{spencer2023kick, spencer2024kick}, as well as to incorporate vision foundation models (VFMs) as backbones \cite{yu2024vfm, spencer2024kick}. Despite these advances, overall performance improvements remain limited. As demonstrated by our preliminary experiments (see Fig.~\ref{fig.teaser}(b)), when existing methods are trained simultaneously on diverse monocular video data (\textit{e.g.}, indoor, outdoor, dynamic, and adverse weather scenarios), they fail to benefit mutually from the increased diversity and do not exhibit improved generalization to out-of-distribution scenes. Instead, we observe a performance decline compared to models trained individually on a single dataset. This observation demonstrates that existing SSMDE methods have not yet achieved effective scalability to diverse data.

This study begins by investigating the challenges that arise when SSMDE is exposed to diverse data distributions during training. It is worth noting that the novel-view synthesis-based supervision inevitably introduces noisy and artifact-laden supervisory signals in scenarios involving occlusions, textureless regions, and illumination changes \cite{wang2025jasmine, guo2026rebenchmarking, liu2026disparity, zhou2026aergs}. Even setting aside such noise, the gradients induced by self-supervised losses often fail to reflect depth errors faithfully \cite{zhang2024dcpi}. As a result, the optimization process is inherently prone to local minima, leading to poor convergence stability \cite{zhang2024dcpi}. Although these challenges frequently occur during self-supervised training, existing methods can still achieve reasonable accuracy on a single, homogeneous dataset \cite{sun2023sc}. This is largely because the network can exploit the structural bias inherent to that dataset, thereby compensating for weak supervision. The key evidence we find is that under the existing SSMDE framework, the coarsest, high-level feature of the depth encoder does not capture depth-related cues, such as the foreground–background relation and perspective geometry. Instead, it primarily reflects the dataset-specific structural biases. When training on diverse datasets, different data domains often require the model to learn inconsistent structural biases for their respective convergence, leading to the observed performance drop.

These findings motivate us to explicitly guide the network to learn a robust, generalized scene-structure representation for depth decoding. Based on this core idea, we develop the \textbf{MaSS} framework that \textbf{Ma}ke \textbf{S}elf-supervised depth learning \textbf{S}calable to diverse data. MaSS bridges a conventional depth encoder with a frozen pretrained VFM through a bidirectional interaction architecture, drawing inspiration from \cite{chen2022vision}. In this design, the depth encoder focuses on extracting geometric cues that are difficult to recover from VFM features \cite{yu2024vfm}. The interacted VFM features are then fed into two feature projectors to reconstruct complementary, collapse-free representations under explicit constraints and learning objectives. Specifically, one projector is adapted to learn a contextual representation informative for depth estimation, such as object-level spatial relationships, while retaining generalized semantics with the original VFM features as anchors. The other projector is then constrained to be decoupled from the contextual one, thereby reserving its representational capacity for scene structure.
To explicitly define this representation, we introduce a scene-structure discrimination objective via contrastive learning. This objective organizes the feature space based on scene layout similarity, which is measured by the histogram distribution of predicted depth. Driven by this objective, the second projector learns transferable structural representations that capture global scene layout and perspective geometry, rather than dataset-specific biases. Together, these two complementary representations enable more reliable depth predictions while reducing their reliance on dataset-specific biases.

Extensive experiments demonstrate that MaSS outperforms previous state-of-the-art (SoTA) self-supervised methods by a large margin (as presented in Fig.~\ref{fig.teaser}(a)) and achieves accuracy comparable to certain supervised depth foundation models. Moreover, MaSS represents the first self-supervised approach to achieve clear performance gains from increasing the diversity of training data (Fig.~\ref{fig.teaser}(b)), resulting in notable generalizability improvements particularly on out-of-distribution scenes (Fig.~\ref{fig.teaser}(c)).

\section{Related Works}

To alleviate the reliance on extensive ground-truth data in monocular depth estimation (MDE), SSMDE from monocular video, first introduced in the study \cite{zhou2017unsupervised}, has attracted growing attention. Within this paradigm, improvements have focused on enhancing photometric supervision \cite{godard2019digging}, addressing scale inconsistency \cite{bian2021unsupervised}, and incorporating additional visual cues to handle occlusions and dynamic scenes \cite{zou2018df, sun2023sc, sun2023dynamo, shao2024monodiffusion, zhang2024dcpi}.
Though significant progress has been made, these SSMDE methods are typically trained on single domains with limited-scale datasets and rely on task-specific network architectures \cite{zhao2022monovit, he2022ra, zhang2023lite, zhou2024recurrent}. The study \cite{gordon2019depth} first demonstrates the feasibility of self-supervised depth learning from heterogeneous in-the-wild videos by jointly estimating depth, camera motion, object motion, and camera intrinsics. With the rise of VFMs \cite{oquab2024dinov, rombach2022high}, increasing efforts have been devoted to developing depth foundation models (DFMs) \cite{yang2024depth, ke2024repurposing, he2025lotus, wang2025editor} trained on large-scale annotated data, achieving strong performance and generalization.
Building upon this trend, recent studies \cite{spencer2023kick, spencer2024kick} have taken initial steps toward scaling SSMDE training, while the others \cite{wang2025jasmine, zhang2025hybrid, zhang2025discriminately} demonstrate that developing DFMs in a self-supervised fashion requires not only harnessing the large pretrained models' capacity, but also addressing challenges unique to self-supervised learning. Among them, cross-domain learning conflicts under mixed-domain training pose a fundamental yet underexplored obstacle to scalability, which this work primarily aims to address.

Three lines of research inform our approach. The first draws on pretrained knowledge to improve depth representations. For example, the study \cite{guizilini2020semantically} uses a pretrained semantic network to guide the learning of geometric features. The studies~\cite{xu2025pixel, simeoni2025dinov3} further emphasize preserving transferable pretrained representations during adaptation. The second focuses on organizing depth representations. The prior arts \cite{wang2024weatherdepth,wang2024digging,gao2024color,cao2025learning,zhang2025hybrid} have introduced contrastive objectives that structure the feature space according to weather variations, color or geometric consistency, historical predictions, and relationships guided by language. The third conditions depth prediction distributions on scene characteristics to account for domain shifts. For example, the study~\cite{fu2024geowizard} conditions these distributions on scene categories. Drawing on these insights, MaSS explicitly decouples contextual and structural representations, preserves contextual relationships derived from VFMs, and adaptively organizes structural features according to similarities in scene layout. Depth prediction is then conditioned on the resulting structural representations to reduce learning conflicts across domains.

\begin{figure*}[t!]
	\begin{center}
		\includegraphics[width=0.99\textwidth]{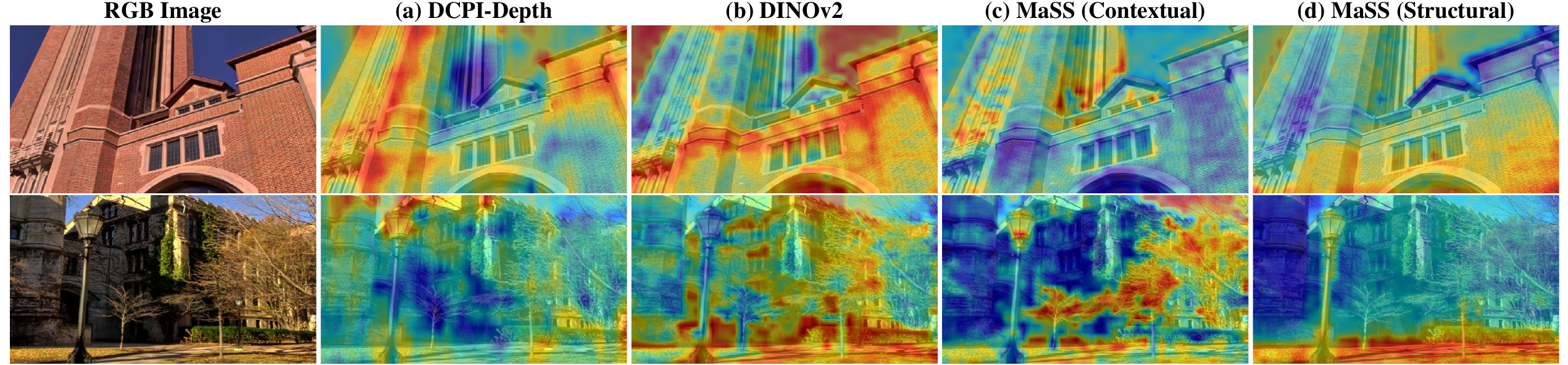}
		\caption{Feature activation map comparison among a conventional self-supervised depth model, DINOv2, and the two representations learned by MaSS. (a) The conventional depth model tends to encode dataset-specific biases, whereas (b) DINOv2 lacks sufficient geometric cues. In contrast, MaSS decouples features into contextual and structural representations: (c) contextual ones capture fine-grained object details and foreground–background relations; (d) structural ones encode global scene layouts that are transferable across diverse datasets.}
		\label{fig.vis_feat}
	\end{center}
\vspace{-1em}
\end{figure*}

\section{Methodology}
\label{sect.3}

SSMDE models are prone to learning dataset-specific inductive biases and struggle to generalize across varying depth distributions during both training and inference. For example, Fig.~\ref{fig.teaser}(c) presents depth predictions on out-of-distribution indoor and outdoor scenes from two representative self-supervised depth models \cite{sun2023sc, zhang2024dcpi}, trained on the NYUv2 \cite{silberman2012indoor} and KITTI \cite{geiger2012we} datasets, respectively. It can be observed that the model trained on driving scenes tends to predict depth with a vanishing point, while the model trained on small rooms fails to capture the more uniform depth distributions in large halls. To find the source of such inductive biases, we investigate the internal feature representations. As shown in Fig.~\ref{fig.vis_feat}(a), the DepthNet encoder \cite{zhang2024dcpi} trained on driving data produces feature activation maps with a consistent low-activation region. This pattern closely resembles the biased depth predictions in Fig.~\ref{fig.teaser}(c) and fails to reflect variations aligned with the actual scene geometry. These results indicate that the dataset-specific inductive biases are already embedded in the intermediate representations. The above findings inspire us with a possible solution to make self-supervised learning scalable: explicitly guide the network to learn robust, generalized structural representations that are discriminative of scene layout before decoding depth.

\begin{figure*}[t!]
	\begin{center}
		\includegraphics[width=0.99\textwidth]{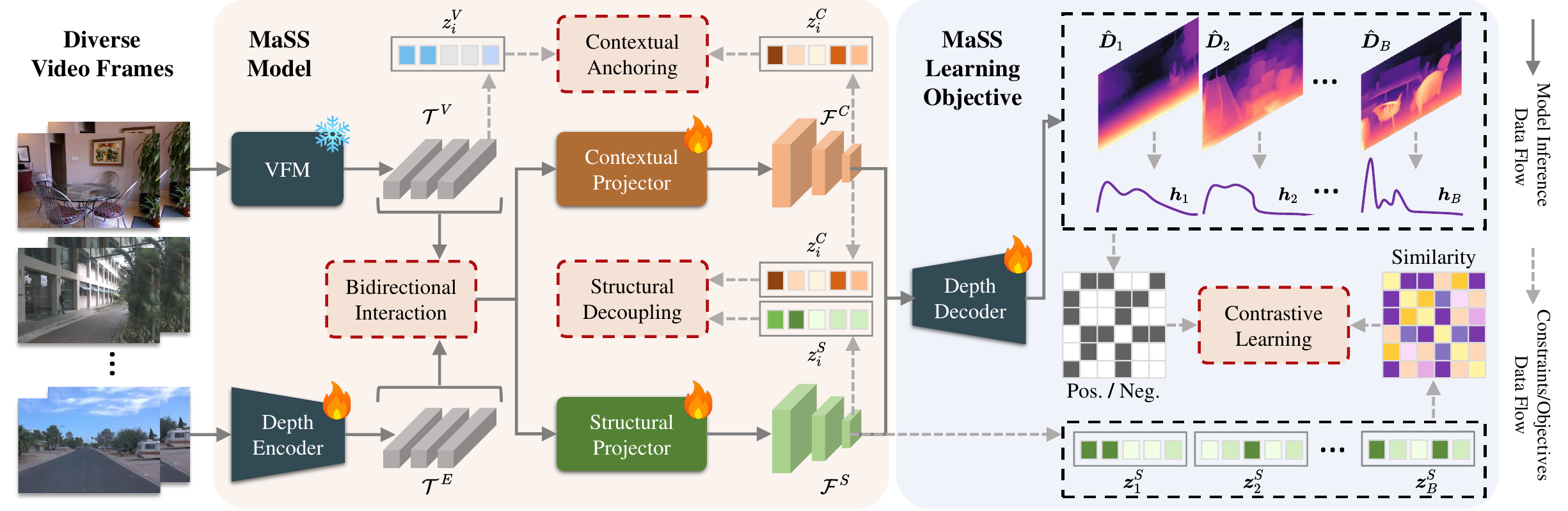}
		\caption{The overall architecture of the proposed MaSS framework.
        MaSS bridges a conventional depth encoder with a frozen pretrained VFM via a bidirectional interaction module. The interacted VFM features are fed into a pair of anchored–decoupled feature projectors, producing collapse-free contextual and structural representations. The structural projectors learn generalized representations from discriminating scene structures, making the self-supervised depth learning scalable.}
		\label{fig.overview}
	\end{center}
\vspace{-0.5em}
\end{figure*}

\subsection{MaSS Model}
\label{sect.3.2}

The first restriction of existing approaches towards this goal is from the model architecture design.
Although the pretrained VFM (\textit{e.g.}, DINOv2 \cite{oquab2024dinov}) exhibits strong generalization and is widely adopted as initialization for supervised depth foundation models \cite{yang2024depth, yang2025depth, wang2025vggt}, its adaptation is unreliable under self-supervision due to noisy training signals \cite{wang2025jasmine}. Direct fine-tuning may deteriorate the pretrained representations \cite{yan2025er}, yet keeping the VFM frozen leaves its semantic features containing less sufficient geometric cues for structural modeling \cite{yu2024vfm} (as seen in Fig.~\ref{fig.vis_feat}(b)). This dilemma motivates a bidirectional interaction architecture between a trainable traditional depth encoder and a frozen VFM encoder, drawing on the ViT-Adapter design \cite{chen2022vision}.
Another key observation is that different viewpoints of the same scene, or different scenes containing similar objects, may appear contextually alike yet differ substantially in spatial structure. Conversely, visually distinct scenes can share similar structural layouts.
Such misalignment suggests that entangling contextual and structural features within a single representation pathway may lead to representational conflict and collapse.
To address this issue, we decouple contextual and structural representations using two specialized feature projectors.
\\
\textbf{Bidirectional interaction architecture.}
As shown in Fig.~\ref{fig.overview}, we first split the ViT blocks of the VFM encoder into three subsets and extract an intermediate token set
 $\mathcal{T}^V = \{\boldsymbol{T}^{V}_i\}_{i=1}^3$,
which consists of $D$-dimensional feature tokens $\boldsymbol{T}^{V}_i \in \mathbb{R}^{\frac{HW}{16^2}\times D}$.
Meanwhile, the depth encoder \cite{zhang2023lite} extracts a multi-scale feature pyramid $\mathcal{F}^E = \{\boldsymbol{F}^E_i\}_{i=1}^3$, corresponding to spatial resolutions of $1/8, 1/16, 1/32$, respectively. We apply $1 \times 1$ convolution to unify the channel dimensions, followed by flattening and concatenation, to obtain $D$-dimensional feature tokens $\boldsymbol{T}^{E}_1 \in \mathbb{R}^{\left(\sum_i\frac{HW}{S_i^2}\right)\times D}$, where $S_i=2^{i+2}$.
The two branches exchange complementary information through the interaction module introduced in \cite{chen2022vision}.
Concretely, the VFM tokens are updated via cross-attention ($\mathrm{Attn}(\cdot)$) with the depth tokens:
\begin{equation}
    \hatsup{\boldsymbol{T}}{V}_i = \boldsymbol{T}^{V}_i + \mathrm{Attn}(\mathrm{LN}(\boldsymbol{T}^{V}_i), \mathrm{LN}(\boldsymbol{T}^{E}_i)),
\end{equation}
where $\mathrm{LN(\cdot)}$ is LayerNorm, while the depth tokens are updated by attending to the VFM tokens, followed by a feed-forward network ($\mathrm{FFN}(\cdot)$):
\begin{equation}
    \hatsup{\boldsymbol{T}}{E}_{i+1} = \hatsup{\boldsymbol{T}}{E}_i+ \mathrm{FFN}(\mathrm{LN(}\hatsup{\boldsymbol{T}}{E}_i)),  \quad \hatsup{\boldsymbol{T}}{E}_i = \boldsymbol{T}^{E}_i + \mathrm{Attn}(\mathrm{LN}({\boldsymbol{T}}^{E}_i), \mathrm{LN}(\boldsymbol{T}^{V}_{i+1}))
\end{equation}
This bidirectional interaction aligns heterogeneous features within a shared latent space while injecting learnable geometric cues from the depth encoder into the frozen VFM representations.
Notably, unlike its common use as an auxiliary module in other fields \cite{xu2023side, guo2026fully}, this architecture plays an indispensable role in enabling structural representation learning and scalable self-supervised training in our work, as demonstrated by the ablation results in Sect.~\ref{sect.abl}.
\\
\textbf{Anchored-decoupled feature projectors.} Next, we introduce two feature projectors that take the interacted VFM features as input and progressively upsample them to produce decoder-compatible feature pyramids $\mathcal{F}^C=\{\boldsymbol{F}_{i}^{C}\}_{i=1}^3$ and $\mathcal{F}^S=\{\boldsymbol{F}_{i}^{S}\}_{i=1}^3$, corresponding to contextual and structural representations, respectively.
Concretely, $\mathcal{T}^V$ are reshaped from sequence form into spatial feature maps $\mathcal{F}^V=\{\hatsup{\boldsymbol{F}}{V}_{i}\}_{i=1}^3$, which are used to compute feature pyramids as
\begin{equation}
\boldsymbol{F}_i^{k}=\mathrm{ReLU}\left(\mathrm{BN}^{k}\left(\mathrm{TransConv}_{2^{i-1}\times2^{i-1}}^{k}\left(\hat{\boldsymbol{F}}_i^{V}\right)\right)\right),\quad k\in\{C,S\}, \ i\in\{1,2,3\}.
\end{equation}
where $\mathrm{TransConv^k}(\cdot)$ denotes transposed convolution, $\mathrm{BN^k}(\cdot)$ represents batch normalization, and $\mathrm{ReLU}(\cdot)$ is the ReLU activation.
To enforce robust and generalizable adaptation of the contextual representation while encouraging the structural pathway to focus on the complementary geometric cues, we introduce an anchoring-decoupling constraint. Specifically, we aggregate the highest-level feature maps $\boldsymbol{F}_{3}^{C}$, $\boldsymbol{F}_{3}^{S}$, and $\boldsymbol{F}_{3}^{V}$ from the VFM into latent descriptors via global average pooling followed by normalization. For a batch of size $B$, this yields three sets of normalized descriptors: $\{\boldsymbol{z}_i^{C}\}_{i=1}^B$, $\{\boldsymbol{z}_i^{S}\}_{i=1}^B$, and $\{\boldsymbol{z}_i^{V}\}_{i=1}^B$.
The introduced constraint can then be formulated as follows:
\begin{equation}
\mathcal{L}_{\text{ac}} =  \sum_{i=1}^B\sum_{j=1, j \ne i }^B({\boldsymbol{z}_i^{C}}^{\top}\boldsymbol{z}_j^{C}  - {\boldsymbol{z}_i^{V}}^{\top}\boldsymbol{z}_j^{V} )^2 + \sum_{i=1}^B ( {\boldsymbol{z}_i^{C}}^{\top}\boldsymbol{z}_i^{S})^2,
\end{equation}
where the first term anchors the contextual features to the semantic relationships captured by the VFM across samples, ensuring that semantic context is preserved during task-specific adaptation. The second term decouples the structural features from contextual ones, encouraging them to focus solely on the complementary geometric cues that reflect scene layout and viewpoint variations, rather than redundant semantic content.

The above two architectural designs jointly establish a geometrically enriched and collapse-free structural pathway for representation learning, for which the corresponding learning objective is introduced in the following section.

\subsection{MaSS Learning Objective}
\label{sect.3.3}

Existing self-supervision signals mainly penalize local photometric inconsistencies \cite{zhang2024dcpi, wang2025jasmine}, which predominantly supervise contextual features and fail to guide the learning of global scene structure. Consequently, an additional objective is required to formulate structural representation learning.
However, directly supervising structural representations is infeasible, as the mapping between intermediate feature distributions and the underlying depth distribution is not analytically tractable. We find an alternative solution drawing inspiration from contrastive learning \cite{radford2021learning, zhang2026oracle}, which organizes the feature space through relational supervision rather than explicit labels.
Unlike prior works \cite{zhang2025hybrid, wang2024weatherdepth, cao2025learning} that construct contrastive pairs in a generic instance-level manner \cite{radford2021learning}, we reformulate a dedicated scene-structure discrimination task, where positive and negative pairs are explicitly defined based on a proposed structural discrepancy metric.
\\
\textbf{Structural discrepancy metric.}
Given a training batch denoted as $\{\boldsymbol{I}_i, \hat{\boldsymbol{D}}_i \}_{i=1}^B$ of size $B$, where $\hat{\boldsymbol{D}}_i \in \mathbb{R}^{H \times W}$ represents the predicted depth map corresponding to image $\boldsymbol{I}_i$, the structural discrepancy metric aims to reflect the similarity between samples in the batch in terms of their global depth distribution.
A straightforward approach is to compare their depth histograms. However, such histogram-based comparisons are highly sensitive to local structural variations and high-frequency artifacts \cite{bhat2022localbins, shao2023iebins}.
To emphasize global distributional differences rather than local details, we first downsample each depth map $\hat{\boldsymbol{D}}_i$ by a factor of $s$ to obtain a coarse proxy map $\boldsymbol{M}_i \in \mathbb{R}^{\frac{H}{s} \times \frac{W}{s}}$, where $s=8$ in our implementation.
We then model the depth distribution of $\boldsymbol{M}_i$ as a discrete probability distribution over $K$ bins, with $\{c_i\}_{i=1}^K$ being the bin centers and $\Delta$ denoting the interval width. $K$ is set to 20.
To further mitigate the effect of high-frequency artifacts, we adopt a soft binning strategy inspired by \cite{yan2025synthetic} when computing the depth histogram.
Instead of hard assignments, each depth value contributes to nearby bins with continuous weights. Concretely, for a depth value $d \in \boldsymbol{M}_i$, its contribution to the $k$-th bin is defined as the following function:
\begin{equation}
\Psi_k(d) = \sigma\left(\left({d - c_k + \Delta/2}\right)/{\tau_h}\right) - \sigma\left(\left(d - c_k - \Delta/2\right)/{\tau_h}\right),
\end{equation}
where $\sigma(\cdot)$ denotes the sigmoid function, and $\tau_h$ is a temperature parameter that controls the smoothness of the bin assignment and suppresses high-frequency perturbations.
Based on this formulation, we define a histogram-like probability vector that summarizes the global depth distribution of the $i$-th sample: $\boldsymbol{h}_i \in \mathbb{R}^K$, obtained by marginalizing over all pixels as:
$\boldsymbol{h}_i = [h_{i,1}, \dots, h_{i,K}]^\top$, where$\quad h_{i,k} = \frac{s^2}{HW} \sum_{\boldsymbol{p}} \Psi_k(\boldsymbol{M}_i(\boldsymbol{p}))$.
By construction, $\boldsymbol{h}_i$ satisfies $\sum_{k=1}^K h_{i,k} \approx 1$.
We then define the structural discrepancy metric as the symmetric Kullback–Leibler (KL) divergence between two given samples $i$ and $j$, which is formulated as follows:
\begin{equation}
\mathcal{D}_{\text{kl}}(\boldsymbol{h}_i || \boldsymbol{h}_j) = \frac{1}{2}(\sum_{k=1}^K h_{i,k} \log \left( {h_{i,k}}/{h_{j,k}} \right) + \sum_{k=1}^K h_{j,k} \log \left( {h_{j,k}}/{h_{i,k}} \right) ).
\end{equation}
\\
\textbf{Learning to discriminate scene structures.}
To explicitly guide the structural feature projector to learn generalized, transferable structural representations, we introduce a scene-structure discrimination objective via contrastive learning.
Specifically, given a sample $i$ within a training batch, the structural descriptor $\boldsymbol{z}_i^S$ is encouraged to be similar to descriptors of samples with similar scene structures, while being dissimilar to those of structurally different samples.
This objective naturally leads to a contrastive learning formulation in the structural representation space. The positive sample set $\mathcal{P}_i$ for anchor $i$ is determined based on the previously defined structural discrepancy metric. We transform the discrepancy into a similarity measure and then threshold it with a hyperparameter $\gamma$:
\begin{equation}
\mathcal{P}_i = \left\{ j \mid \exp\left( - \mathcal{D}_{\text{kl}}(\boldsymbol{h}_i | \boldsymbol{h}_j) \right) > \gamma \right\},
\end{equation}
where we set $\gamma$ to 0.95 based on empirical tuning.
Following common practice in contrastive learning \cite{radford2021learning,zhang2025hybrid}, each sample is also treated as a positive to itself. Based on the resulting positive set, we adopt a multi-positive InfoNCE objective to optimize the structural representations, formulated as:
\begin{equation}\mathcal{L}_{\textbf{scl}} = - \frac{1}{B}\sum_{i=1}^B \log ({\sum_{j \in \mathcal{P}_i} \exp( {\boldsymbol{z}_i^S}^{\top} \boldsymbol{z}_j^S / \tau_{c})}/{\sum_{k=1}^B \exp({\boldsymbol{z}_i^S}^{\top} \boldsymbol{z}_k^S / \tau_{c})}),\end{equation}
where $\tau_c$ is the contrastive temperature controlling the concentration of the embedding space, and is set to 0.1 in our implementation.
We provide several quantitative examples in the \textit{Supplementary Materials} to demonstrate that the selected hyperparameters effectively prevent this approach from degenerating into discrimination based on coarse scene categories (\textit{e.g.}, indoor and outdoor).

\subsection{Depth Prediction from Learned Representations}
The contextual and structural representations learned by MaSS are visualized in Fig.~\ref{fig.vis_feat}(c)(d). The contextual pathway effectively adapts the VFM representations into depth-related context, while the structural pathway captures generalized scene geometry. Their corresponding feature pyramids are then concatenated along the channel dimension and passed through a $1\times1$ convolution for dimension reduction, and finally fed into the depth decoder.

\section{Experiments}

\subsection{Experimental Setups}

\textbf{Implementation details.}
The MaSS model utilizes the pre-trained DINOv2 ViT-B/14 architecture \cite{oquab2024dinov} as its VFM encoder and employs the depth encoder \cite{zhang2023lite}.
The interaction module follows the design of ViT‑Adapter \cite{chen2022vision}. During training, the DINOv2 encoder remains frozen while all other parameters are updated.
The MaSS model can be trained on a single RTX 4090 GPU with a batch size of 6, which takes approximately 2.5 days to complete.
Data augmentation strategy and optimizer settings adhere to the previous studies \cite{godard2019digging, zhang2023lite, zhang2024dcpi} without additional tuning.
The proposed scene-structure discrimination objective operates on the depth predictions to formulate supervision for representation learning.
Consequently, it is infeasible to train the depth model from scratch with this objective, as unreliable early predictions would adversely affect representation learning. Therefore, an additional initialization stage for one epoch is required to obtain reasonable depth estimates before applying this objective.
During training, we follow the prior work~\cite{sun2023sc} to handle dynamic scenes and extreme conditions that may compromise training stability~\cite{yan2025synthetic, zhang2025discriminately}.
\\
\textbf{Datasets.}
Experiments are conducted on a total of 13 public datasets. The KITTI \cite{geiger2012we}, NYUv2 \cite{silberman2012indoor}, DDAD \cite{guizilini20203d}, and nuScenes \cite{caesar2020nuscenes} datasets are used for both training and evaluation, covering outdoor, indoor, dynamic, and adverse weather scenarios.
Eight additional datasets are used to evaluate the generalizability of MaSS to unseen data, including three driving datasets, Waymo Open \cite{mei2022waymo}, DrivingStereo \cite{yang2019drivingstereo}, and Cityscapes \cite{cordts2016cityscapes}; three indoor datasets, iBims-1 \cite{koch2020ibims}, SUN-RGBD \cite{song2015sun}, and 7-Scenes \cite{shotton20137scene}; and two out-of-distribution datasets, DIML \cite{cho2021diml} and DIODE \cite{vasiljevic2019diode}.
Finally, the TartanAir dataset is used to expand the size and diversity of the training data to assess the scalability of MaSS.
The characteristics of each dataset and the adopted data splits, following those established in prior works \cite{zhou2017unsupervised, sun2023dynamo, wang2025jasmine, bian2021auto, sun2023sc, zhang2026midas}, are detailed in the \textit{Supplementary Materials}.
\\
\textbf{Evaluation protocol.}
In addition to self-supervised methods that inherently predict scale-invariant (SI) depth, we also include several supervised depth foundation models that produce affine-invariant (SSI) predictions for comparison. For scale-invariant evaluation, predicted depths are aligned to ground truth using median scaling (Median). For affine-invariant evaluation, predictions are aligned to ground-truth depth via least-squares fitting (LSQ).
We report two widely used metrics: Absolute Mean Relative (Abs Rel) error and $\delta_1$ accuracy.

\subsection{Compare with SoTA Self-supervised Methods}

\begin{table}[t]
\centering
\settablefont
\caption{Quantitative results on the training-evaluation datasets.
$^{\dagger}$ indicates training under the mixed-dataset setting. For single-dataset training, results are reported at the image resolutions of 192$\times$640, 384$\times$640, 256$\times$320, and 288$\times$512, respectively, while mixed-dataset training results are evaluated at a unified resolution of 384$\times$640. The best results are marked in bold type. }
\label{tb.train}
\begin{tabular}{lllcccccccc} %
\toprule
\multirow{2}{*}[-0.75ex]{Method} & \multirow{2}{*}{Type} & \multirow{2}{*}{Align.} & \multicolumn{2}{c}{KITTI} & \multicolumn{2}{c}{DDAD} &
\multicolumn{2}{c}{NYUv2} & \multicolumn{2}{c}{nuScenes} \\
\cmidrule(lr){4-5}\cmidrule(lr){6-7}\cmidrule(lr){8-9}\cmidrule(lr){10-11} %
&&& Abs~Rel $\downarrow$ & $\delta_1$ $\uparrow$ & Abs~Rel $\downarrow$ & $\delta_1$ $\uparrow$
& Abs~Rel $\downarrow$ & $\delta_1$ $\uparrow$ & Abs~Rel $\downarrow$ & $\delta_1$ $\uparrow$ \\
\midrule
SC-DepthV2~\cite{bian2021auto} \tiny\clgray{TPAMI'22} & SI & Median & 0.118 & 0.866 & 0.164 & 0.775 & 0.138 & 0.820 & -- & -- \\
Lite-Mono~\cite{zhang2023lite} \tiny\clgray{CVPR'23} & SI & Median & 0.101 & 0.897 & 0.175 & 0.799 & 0.198 & 0.688 & 0.419 & 0.720 \\
Dynamo-Depth~\cite{sun2023dynamo} \tiny\clgray{NeurIPS'23} & SI & Median & 0.112 & 0.873 & 0.150 & 0.798 & -- & -- & 0.179 & 0.787 \\
Gas-Mono~\cite{zhao2023gasmono} \tiny\clgray{ICCV'23} & SI & Median & 0.111 & 0.894 & 0.178 & 0.756 & 0.113 & 0.871 & -- & -- \\
SC-DepthV3~\cite{sun2023sc} \tiny\clgray{TPAMI'24} & SI & Median & 0.118 & 0.864 & 0.142 & 0.813 & 0.123 & 0.848 & 0.167 & 0.784 \\
DCPI-Depth~\cite{zhang2024dcpi} \tiny\clgray{TIP'25} & SI & Median & 0.095 & 0.902 & 0.140 & 0.815 & 0.151 & 0.785 & 0.160 & 0.793 \\
GeoDepth~\cite{wu2025geodepth} \tiny\clgray{CVPR'25} & SI & Median & 0.100 & 0.897 & -- & -- & 0.134 & 0.833 & -- & -- \\
Jasmine~\cite{wang2025jasmine} \tiny\clgray{NeurIPS'25} & SI & Median & 0.097 & 0.907 & -- & -- & -- & -- & -- & -- \\
DiMoDE~\cite{zhang2025discriminately} \tiny\clgray{TPAMI'26} & SI & Median & 0.097 & 0.899 & 0.134 & 0.831 & -- & -- & 0.139 & 0.836 \\
\textbf{MaSS (Ours)} & SI & Median & \textbf{0.094} & \textbf{0.909} & \textbf{0.128} & \textbf{0.845} & \textbf{0.109} & \textbf{0.868} & \textbf{0.139} & \textbf{0.851} \\
\midrule
Lite-Mono$^\dagger$~\cite{zhang2023lite} \tiny\clgray{CVPR'23} & SI & Median & 0.111 & 0.886 & 0.202 & 0.758 & 0.198 & 0.672 & 1.147 & 0.816 \\
Dynamo-Depth$^\dagger$~\cite{sun2023dynamo} \tiny\clgray{NeurIPS'23} & SI & Median & 0.126 & 0.844 & 0.165 & 0.764 & 0.312 & 0.479 & 0.181 & 0.710 \\
SC-DepthV3$^\dagger$~\cite{sun2023sc} \tiny\clgray{TPAMI'24} & SI & Median & 0.132 & 0.832 & 0.158 & 0.795 & 0.156 & 0.773 & 0.226 & 0.685 \\
DCPI-Depth$^\dagger$~\cite{zhang2024dcpi} \tiny\clgray{TIP'25} & SI & Median & 0.114 & 0.870 & 0.162 & 0.777 & 0.331 & 0.472 & 0.160 & 0.776 \\
\textbf{MaSS$^\dagger$ (Ours)} & SI & Median & \textbf{0.092} & \textbf{0.913} & \textbf{0.125} & \textbf{0.862} & \textbf{0.104} & \textbf{0.890} & \textbf{0.134} & \textbf{0.852} \\
\bottomrule
\end{tabular}
\vspace{-0.5em}
\end{table}

\begin{figure*}[t!]
	\centering
	\includegraphics[width=0.99\textwidth]{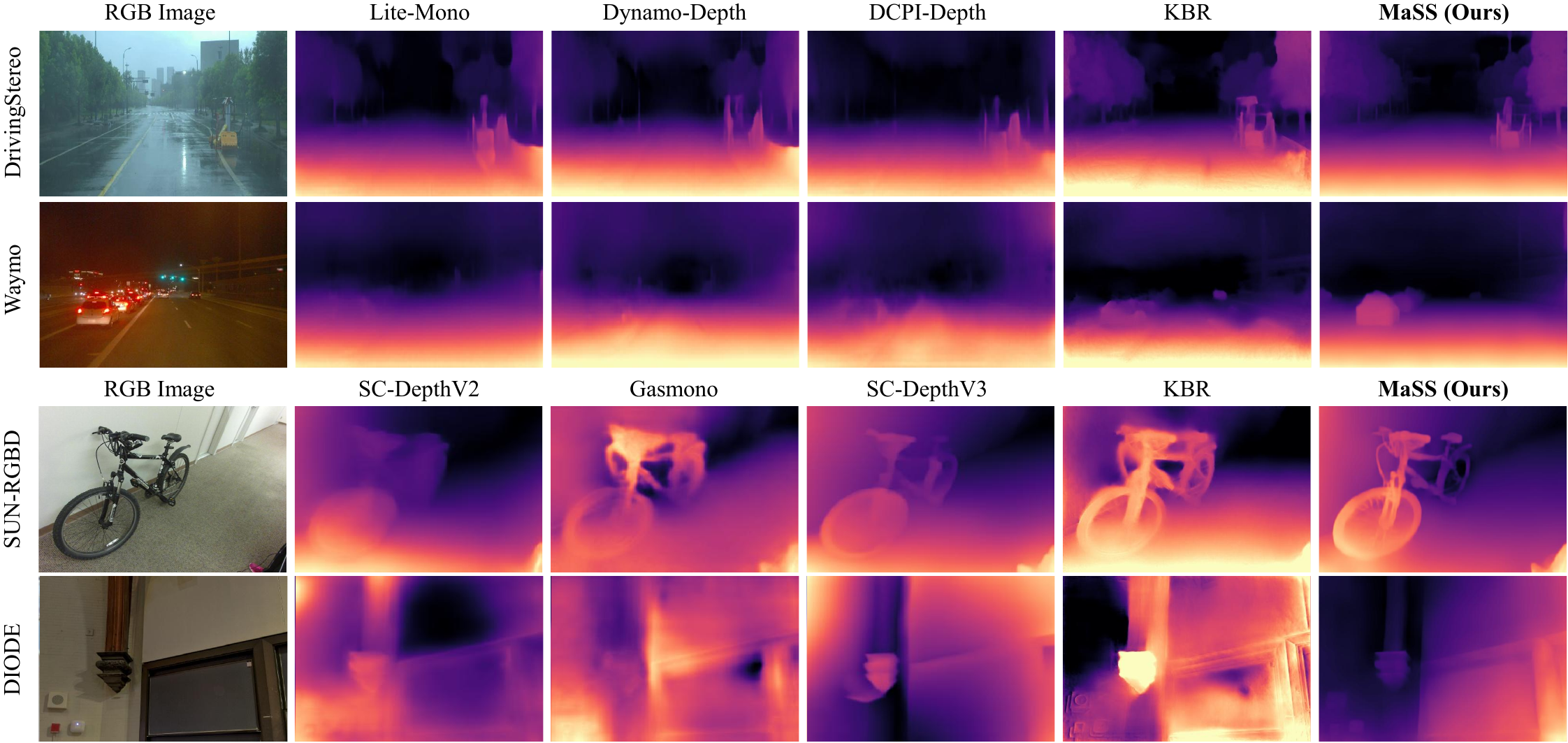}
\caption{Qualitative results on zero-shot generalization datasets. Compared with representative SSMDE models specialized for either indoor or outdoor scenes and the model scaled with large-scale data and pretrained backbones, MaSS demonstrates consistent superiority across both scenarios.
}
\label{fig.outdepth}
\vspace{-1em}
\end{figure*}

\textbf{Individual dataset training.}
Table \ref{tb.train} shows that MaSS outperforms all previous SoTA methods \cite{bian2021auto, sun2023dynamo, zhao2023gasmono, sun2023sc, wu2025geodepth, zhang2024dcpi, wang2025jasmine} across all four datasets when trained and evaluated individually.
\\
\textbf{Mixed dataset training.}
While our models achieve strong results when trained on individual datasets, such settings do not reflect the scalability challenges posed by large-scale and diverse training data, which constitute the primary focus of this work.
To this end, we further conduct comparative experiments in a mixed-dataset training setting. Specifically, each training batch is assembled by randomly sampling data from four datasets, thereby simulating a realistic, large-scale training scenario with substantial data diversity. To maintain a comparable number of training iterations to the individual-dataset setting, the number of training epochs is reduced to one-third. Under this setting, we reproduce several SoTA methods \cite{zhang2023lite,sun2023dynamo,sun2023sc,zhang2024dcpi} with publicly available training code and compare them with our approach.
As shown in Table \ref{tb.train}, although the compared SoTA methods are robust across different datasets in individual dataset training, meaning no single dataset causes training to fail, their performance still drops significantly in mixed dataset training.
In contrast, our method consistently achieves clear performance improvements.

\begin{table}[t]
\begin{center}
\settablefont
\caption{
Quantitative zero-shot generalization results.
Models under the single dataset training are evaluated at the image resolutions of 192$\times$640 (outdoor) and 256$\times$320 (indoor). Mixed-dataset models ($^{\dagger}$) are evaluated for the image resolutions of 384$\times$640. Supervised foundation models are reported at their pretrained resolutions (518$\times$518 for \cite{yang2025depth} and 768$\times$768 for \cite{ke2024repurposing, he2025lotus}). Results on the DIML \cite{cho2021diml} dataset are not reported for method \cite{yang2025depth} that was pretrained on this dataset. The best results for self-supervised and supervised models are marked in bold type, respectively. SSI* and LSQ* denote the predicted depth and its alignment with the ground-truth in the inverse depth domain.}
\label{tb.zeroshot}
\begin{tabular}{lllcccccccc}
\toprule
\multirow{2}{*}[-0.75ex]{Method} & \multirow{2}{*}{Type}
& \multirow{2}{*}{Align.} & \multicolumn{2}{c}{Waymo} & \multicolumn{2}{c}{DrivingStereo} &
\multicolumn{2}{c}{Cityscapes} & \multicolumn{2}{c}{DIML} \\
\cmidrule(lr){4-5}\cmidrule(lr){6-7}\cmidrule(lr){8-9}\cmidrule(lr){10-11}
&&& Abs~Rel $\downarrow$ & $\delta_1$ $\uparrow$ & Abs~Rel $\downarrow$ & $\delta_1$ $\uparrow$
& Abs~Rel $\downarrow$ & $\delta_1$ $\uparrow$ & Abs~Rel $\downarrow$ & $\delta_1$ $\uparrow$ \\
\midrule
DA~V2~\cite{yang2025depth} \tiny\clgray{NeurIPS'24}        &SSI* & LSQ* &  \textbf{0.084} & \textbf{0.912} & \textbf{0.103} & \textbf{0.908} & \textbf{0.088} & \textbf{0.920} & -- & --  \\
Marigold~\cite{ke2024repurposing} \tiny\clgray{CVPR'24}        &SSI & LSQ &  0.128 & 0.868 & 0.128 & 0.858 & 0.118 & 0.886 & 0.123  & 0.863 \\
Lotus~\cite{he2025lotus} \tiny\clgray{ICLR'25}        &SSI & LSQ &  0.123 & 0.881 & 0.123 & 0.877 & 0.121 & 0.878 & \textbf{0.099} & \textbf{0.908} \\
\midrule
Dynamo-Depth~\cite{sun2023dynamo} \tiny\clgray{NeurIPS'23}        & SI & Median & 0.167 & 0.784 & 0.164 & 0.788 & 0.233 & 0.564 & 0.175 & 0.748  \\
DCPI-Depth~\cite{zhang2024dcpi} \tiny\clgray{TIP'25}        & SI & Median & 0.178 & 0.771 & 0.161 & 0.787 & 0.241 & 0.549 & 0.174 & 0.733  \\
Jasmine~\cite{wang2025jasmine} \tiny\clgray{NeurIPS'25}        & SI & Median & 0.124 & 0.864 & 0.148 & 0.826 & 0.119 & 0.855 & 0.199 & 0.739 \\
\textbf{MaSS (Ours, KITTI)}        & SI & Median & 0.141 & 0.836 & 0.131 & 0.850 & 0.132 & 0.846 & 0.115 & 0.875  \\
\hdashline
KBR~\cite{spencer2023kick} \tiny\clgray{ICCV'23} & SI & Median &  0.156 & 0.806 & 0.139 & 0.825 & 0.111 & 0.864 & 0.105 & \textbf{0.902}  \\
KBR$++$~\cite{spencer2024kick} \tiny\clgray{Arxiv'24} & SI & Median & 0.160 & 0.803 & 0.126 & 0.863 & 0.105 & 0.882 & 0.108 & 0.901  \\
\textbf{MaSS$^\dagger$ (Ours)}        &SI & Median &  \textbf{0.123} & \textbf{0.867} & \textbf{0.100} & \textbf{0.905} & \textbf{0.103} & \textbf{0.896} & \textbf{0.104} & 0.890  \\
\midrule
\midrule
\multirow{2}{*}[-0.75ex]{Method} & \multirow{2}{*}{Type}
& \multirow{2}{*}{Align.} & \multicolumn{2}{c}{iBims} & \multicolumn{2}{c}{SUN-RGBD} &
\multicolumn{2}{c}{7-Scenes} & \multicolumn{2}{c}{DIODE} \\
\cmidrule(lr){4-5}\cmidrule(lr){6-7}\cmidrule(lr){8-9}\cmidrule(lr){10-11}
&&& Abs~Rel $\downarrow$ & $\delta_1$ $\uparrow$ & Abs~Rel $\downarrow$ & $\delta_1$ $\uparrow$
& Abs~Rel $\downarrow$ & $\delta_1$ $\uparrow$ & Abs~Rel $\downarrow$ & $\delta_1$ $\uparrow$ \\
\midrule
DA~V2~\cite{yang2025depth} \tiny\clgray{NeurIPS'24}        & SSI* & LSQ* & \textbf{0.045} & \textbf{0.981} & \textbf{0.062} & \textbf{0.961} & \textbf{0.074} & \textbf{0.940} & 0.265 & 0.734  \\
Marigold~\cite{ke2024repurposing} \tiny\clgray{CVPR'24}        & SSI & LSQ & 0.060 & 0.958 & 0.081 & 0.943 & 0.086 & 0.921 & 0.308 & \textbf{0.773} \\
Lotus~\cite{he2025lotus} \tiny\clgray{ICLR'25}        & SSI & LSQ & 0.053 & 0.969 & 0.080 & 0.951 & 0.082 & 0.929 & \textbf{0.229} & 0.729  \\
\midrule
Gas-Mono~\cite{zhao2023gasmono} \tiny\clgray{ICCV'23}        & SI & Median & 0.153 & 0.775 & 0.161 & 0.795 & 0.137 & 0.825 & 0.369 & 0.624  \\
SC-DepthV3~\cite{sun2023sc} \tiny\clgray{TPAMI'24}        & SI & Median & 0.136 & 0.812 & 0.150 & 0.822 & 0.140 & 0.799 & 0.405 & 0.637  \\
\textbf{MaSS (Ours, NYUv2)}        & SI & Median & 0.113 & 0.869 & 0.139 & 0.865 & 0.144 & 0.792 & 0.312 & 0.711  \\
\hdashline
KBR~\cite{spencer2023kick} \tiny\clgray{ICCV'23} & SI & Median & 0.139 & 0.828 & 0.163 & 0.816 & 0.170 & 0.758 & 0.328 & 0.665  \\
KBR$++$~\cite{spencer2024kick} \tiny\clgray{Arxiv'24} & SI & Median & 0.121 & 0.871 & 0.154 & 0.840 & 0.179 & 0.763 & 0.431 & 0.627  \\
\textbf{MaSS$^\dagger$ (Ours)}        & SI & Median & \textbf{0.107} & \textbf{0.883} & \textbf{0.127} & \textbf{0.893} & \textbf{0.120} & \textbf{0.861} & \textbf{0.289} & \textbf{0.736}  \\
\bottomrule
\end{tabular}
\end{center}
\vspace{-1.5em}
\end{table}

\subsection{Zero-Shot Generalizability Evaluation}

\begin{table}[t]
\begin{center}
\setlength{\tabcolsep}{3pt}
\settablefont
\caption{Quantitative comparisons on the models' overhead and their required training data.}
\label{tb.efficiency}
\begin{tabular}{lccccccccc}
\toprule
Method
& Marigold \cite{ke2024repurposing} & DA V2 \cite{yang2025depth} & Lite-Mono \cite{zhang2023lite} & DCPI-Depth \cite{zhang2024dcpi} & SC-DepthV3 \cite{sun2023sc} & Jasmine \cite{wang2025jasmine} & KBR \cite{spencer2023kick} & KBR$++$ \cite{spencer2024kick} & \textbf{MaSS$^\dagger$ (Ours)} \\
\midrule
Params. (M) & 1290.0 & 97.5 & 3.1 & 8.8 & 14.8 & 1290.0 & 92.5 & 266.6 & 93.2 \\
FLOPs (G) & 339.3 & 89.5 & 6.2 & 13.7 & 9.8 & 339.3 & 88.7 & 381.7 & 81.6 \\
Training Data & 74K & 62M & 40K & 40K & 20K & 68K & 1.9M & 2.2M & 150K \\
\bottomrule
\end{tabular}
\end{center}
\vspace{-1.5em}
\end{table}

\begin{table}[t]
\begin{center}
\settablefont
\caption{Quantitative results for ablation studies and model design analyses.
}
\label{tb.abl}
\begin{tabular}{lcccccccc}
\toprule
\multirow{3}{*}[-0.75ex]{(ID) Method} & \multicolumn{4}{c}{Training Sets} & \multicolumn{4}{c}{Zero-Shot} \\
\cmidrule(lr){2-5}\cmidrule(lr){6-9}
& \multicolumn{2}{c}{nuScenes} & \multicolumn{2}{c}{NYUv2} &
\multicolumn{2}{c}{Waymo} & \multicolumn{2}{c}{7-Scenes} \\
\cmidrule(lr){2-3}\cmidrule(lr){4-5}\cmidrule(lr){6-7}\cmidrule(lr){8-9}
& Abs~Rel $\downarrow$ & $\delta_1$ $\uparrow$ & Abs~Rel $\downarrow$ & $\delta_1$ $\uparrow$
& Abs~Rel $\downarrow$ & $\delta_1$ $\uparrow$ & Abs~Rel $\downarrow$ & $\delta_1$ $\uparrow$ \\
\midrule
MaSS$^\dagger$       & 0.134 & 0.852 & 0.104 & 0.890 & 0.123 & 0.867 & 0.120 & 0.855  \\
\midrule
(a1) w/o Contrastive Learning         & 0.145 & 0.850 & 0.106 & 0.891 & 0.163 & 0.805 & 0.135 & 0.820  \\
(a2) w/o Representation Decoupling         & 0.144 & 0.840 & 0.116 & 0.862 & 0.148 & 0.831 & 0.128 & 0.823  \\
(a3) w/o Interaction Architecture         & 0.147 & 0.837 & 0.119 & 0.859 & 0.190 & 0.750 & 0.139 & 0.805  \\
\midrule
(b1) DINOv2 \cite{oquab2024dinov} (Frozen) $+$ DPT \cite{ranftl2021vision}      & 0.186 & 0.719 & 0.129 & 0.832 & 0.163 & 0.757 & 0.126 & 0.839  \\
(b2) DINOv2 \cite{oquab2024dinov} (FFT) $+$ DPT \cite{ranftl2021vision}        & 0.197 & 0.709 & 0.167 & 0.747 & 0.226 & 0.681 & 0.215 & 0.600  \\
(b3) ViT-Adapter \cite{chen2022vision} $+$ DPT \cite{ranftl2021vision}      & 0.166 & 0.767 & 0.123 & 0.856 & 0.147 & 0.822 & 0.163 & 0.757  \\
\midrule
(c1) w/ DINOv3 \cite{simeoni2025dinov3}       & 0.136 & 0.855 & 0.100 & 0.897 & 0.152 & 0.826 & 0.128 & 0.834  \\
\bottomrule
\end{tabular}
\end{center}
\vspace{-1em}
\end{table}

Unlike prior works \cite{zhang2023lite, sun2023sc, zhang2024dcpi} that often treat generalizability evaluation as merely supplementary to in-domain results, we conduct a comprehensive zero-shot generalization study, as shown in Table~\ref{tb.zeroshot}.
First, to validate the advantage of learning generalizable structural representations, we compare MaSS with representative outdoor and indoor SSMDE methods \cite{zhang2023lite, sun2023dynamo, zhang2024dcpi, wang2025jasmine, bian2021auto, zhao2023gasmono, sun2023sc}. All models are trained on KITTI for outdoor evaluation and on NYUv2 for indoor evaluation.
Second, to demonstrate the benefit of scaling training with diverse data, we further include our mixed-dataset trained model (MaSS$^{\dagger}$) in comparisons across all zero-shot evaluation datasets.
Furthermore, we include the prior works that directly train SSMDE on large-scale data \cite{spencer2023kick, spencer2024kick}, as well as several supervised depth foundation models \cite{yang2025depth, ke2024repurposing, he2025lotus} in the generalization comparison.
Finally, the comparisons of the models’ overhead and their required training data are included for a more comprehensive discussion on their achieved performance, as provided in Table~\ref {tb.efficiency}.

First, even on datasets \cite{cordts2016cityscapes, koch2020ibims} that are fairly similar to the training data, previous approaches do not necessarily achieve satisfactory performance. In contrast, our method maintains strong generalization across all zero-shot datasets, including those with more complex and challenging environments \cite{yang2019drivingstereo, mei2022waymo, song2015sun}, and especially on out-of-distribution scenes \cite{cho2021diml, vasiljevic2019diode}. This result highlights the advantage of MaSS in learning transferable structural representations rather than relying on dataset-specific biases.
Moreover, MaSS$^{\dagger}$ consistently outperforms MaSS trained on individual datasets, indicating that our method enables mutual enhancement across diverse training data through learning to discriminate scene structure, demonstrating the effectiveness of scalable self-supervised depth learning.
Furthermore, MaSS$^{\dagger}$ outperforms the KBR series \cite{spencer2023kick, spencer2024kick} while requiring less overhead and significantly less training data, highlighting that our method delivers substantial gains through improved learning scalability rather than merely increasing data scale.
Finally, MaSS$^{\dagger}$ achieves competitive performance compared to supervised depth foundation models on several outdoor and out-of-distribution datasets. It is worth noting that MaSS still underperforms supervised models on standard indoor datasets. This is likely because these datasets have relatively simple and homogeneous depth distributions, making them particularly favorable for supervised models trained on large-scale indoor ground-truth data (\textit{e.g.}, Hypersim \cite{roberts2021hypersim} and NYUv2).
This analysis is supported by results on DIODE and DIML, where the performance gap becomes much smaller on indoor scenes with broader depth ranges and on outdoor scenes that exhibit indoor-style depth distributions.

Notably, part of the performance gap compared to DA v2~\cite{yang2025depth}, as well as the inconsistencies with the originally reported results of the KBR series\cite{spencer2023kick, spencer2024kick}, stem from differences in alignment between Median and SSI*. Additional results and analysis of this issue are provided in the \textit{Supplementary Material}.

Qualitative results are presented in Figs.~\ref{fig.outdepth}. Compared with previous methods, MaSS produces sharper object boundaries and finer-grained details (row 3). It also remains robust in complex environments and preserves more accurate scene geometry in these unseen scenarios (rows 1, 2, 4), benefiting from the explicitly learned structural representation.

\subsection{Ablation Study}
\label{sect.abl}

\textbf{Efficacy of each innovative component.}
To verify that the three key components of MaSS for learning generalizable structural representations are all indispensable, we conduct ablation studies in Table~\ref{tb.abl}(a).
Removing (a1) the contrastive learning objective eliminates explicit supervision for structural representation learning, leading to a significant performance drop, particularly on zero-shot datasets.
Removing (a2) the specialized structural pathway for representation decoupling prevents the model from preserving contextual information while learning structure, resulting in consistent degradation across all datasets.
Removing (a3) the interaction architecture forces the model to learn structural representations directly from DINOv2 features, which lack sufficient geometric cues, and thus causes a substantial performance decline.
\\
\textbf{Superiority over alternative designs.}
MaSS builds upon well-established components such as DINOv2 \cite{oquab2024dinov} and ViT-Adapter \cite{chen2022vision}, making it important to verify that the superior performance stems from our design rather than from simply adopting these components. To this end, we compare MaSS with several plausible alternative configurations in Table~\ref{tb.abl}(b):
(b1) directly fine-tuning a DPT \cite{ranftl2021vision} head with a frozen pretrained DINOv2 backbone,
(b2) fully fine-tuning the entire DINOv2–DPT architecture, and
(b3) incorporating ViT-Adapter while fine-tuning both the adapter and the DPT head.
Results show that MaSS outperforms all these alternative designs by a large margin. This confirms our analysis that effective scaling of self-supervised depth learning requires explicitly modeling scene structure, rather than relying solely on stronger pretrained backbones.
\\
\textbf{Effect of VFM backbone.}
As demonstrated in Table~\ref{tb.abl}(c), replacing the VFM backbone with a stronger self-supervised pretrained model (DINOv3 \cite{simeoni2025dinov3}) leads to only marginal performance changes. This result corroborates our earlier analysis that pretrained VFMs in our framework mainly provide generalizable semantic priors and are not directly responsible for depth decoding.
This naturally raises a follow-up question: since pretrained depth or geometry foundation models encode rich geometric priors~\cite{wang2026geometry}, can MaSS benefit from them, or even further enhance such models?
We investigate this question in the \textit{Supplementary Material}.

\begin{table}[t]
\begin{center}
\settablefont
\caption{Quantitative results for the training scalability analyses.}
\label{tb.scale}
\begin{tabular}{lcccccccc}
\toprule
\multirow{3}{*}{Method} & \multicolumn{2}{c}{KITTI} & \multicolumn{2}{c}{DDAD} &
\multicolumn{2}{c}{NYUv2} & \multicolumn{2}{c}{nuScenes} \\
\cmidrule(lr){2-3}\cmidrule(lr){4-5}\cmidrule(lr){6-7}\cmidrule(lr){8-9}
& Abs~Rel $\downarrow$ & $\delta_1$ $\uparrow$ & Abs~Rel $\downarrow$ & $\delta_1$ $\uparrow$
& Abs~Rel $\downarrow$ & $\delta_1$ $\uparrow$ & Abs~Rel $\downarrow$ & $\delta_1$ $\uparrow$ \\
\midrule
MaSS$^\dagger$ (K+D+NY+NU, 150K) & \textbf{0.092} & \textbf{0.913} & 0.125 & \textbf{0.862} & 0.104 & 0.890 & \textbf{0.134} & \textbf{0.852} \\
MaSS$^\dagger$*  (K+D+NY+NU+T, 200K)     & 0.093 & 0.912 & \textbf{0.122} & 0.857 & \textbf{0.097} & \textbf{0.905} & 0.136 & 0.850  \\
\midrule
\midrule
\multirow{3}{*}{Method} & \multicolumn{2}{c}{Waymo} & \multicolumn{2}{c}{DrivingStereo} &
\multicolumn{2}{c}{Cityscapes} & \multicolumn{2}{c}{DIML} \\
\cmidrule(lr){2-3}\cmidrule(lr){4-5}\cmidrule(lr){6-7}\cmidrule(lr){8-9}
& Abs~Rel $\downarrow$ & $\delta_1$ $\uparrow$ & Abs~Rel $\downarrow$ & $\delta_1$ $\uparrow$
& Abs~Rel $\downarrow$ & $\delta_1$ $\uparrow$ & Abs~Rel $\downarrow$ & $\delta_1$ $\uparrow$ \\
\midrule
MaSS$^\dagger$ (K+D+NY+NU, 150K)      & 0.123 & 0.867 & 0.100 & 0.905 & \textbf{0.103} & \textbf{0.896} & 0.104 & 0.890  \\
MaSS$^\dagger$*  (K+D+NY+NU+T, 200K)     & \textbf{0.119} & \textbf{0.878} & \textbf{0.097} & \textbf{0.908} & 0.106 & 0.892 & \textbf{0.087} & \textbf{0.929}  \\
\midrule
\midrule
\multirow{3}{*}{Method} & \multicolumn{2}{c}{iBims} & \multicolumn{2}{c}{SUN-RGBD} &
\multicolumn{2}{c}{7-Scenes} & \multicolumn{2}{c}{DIODE} \\
\cmidrule(lr){2-3}\cmidrule(lr){4-5}\cmidrule(lr){6-7}\cmidrule(lr){8-9}
& Abs~Rel $\downarrow$ & $\delta_1$ $\uparrow$ & Abs~Rel $\downarrow$ & $\delta_1$ $\uparrow$
& Abs~Rel $\downarrow$ & $\delta_1$ $\uparrow$ & Abs~Rel $\downarrow$ & $\delta_1$ $\uparrow$ \\
\midrule
MaSS$^\dagger$ (K+D+NY+NU, 150K)      & 0.107 & 0.883 & \textbf{0.127} & \textbf{0.893} & \textbf{0.120} & \textbf{0.861} & 0.289 & 0.736  \\
MaSS$^\dagger$*  (K+D+NY+NU+T, 200K)     & \textbf{0.105} & \textbf{0.886} & 0.129 & 0.891 & 0.126 & 0.840 & \textbf{0.289} & \textbf{0.746}  \\
\bottomrule
\end{tabular}
\end{center}
\vspace{-1em}
\end{table}

\subsection{Scalability Analyses}
\label{sect.scale_analyses}
To further evaluate the scalability of MaSS as training data grow in size and diversity, 50K additional samples from five representative scenes in the TartanAir dataset: carwelding, hospital, neighborhood, office, and seasonforest. This expands the training set from 150K to 200K samples and introduces a broader range of scene types. As shown in Table~\ref{tb.scale}, the average performance changes in Abs Rel and $\delta_1$ are $+1.5\%$ and $+0.2\%$ on training domains, $-0.3\%$ and $-0.2\%$ on unseen domains similar to the training domains, and $+4.3\%$ and $+3.0\%$ on more challenging out-of-distribution domains (DIML and DIODE datasets). These results suggest that MaSS benefits from larger and more diverse training data, with performance on the original training domains remaining largely stable and generalization improving significantly on the most challenging unseen domains.
Notably, these results provide only preliminary evidence of the scalability of MaSS, as our evaluation remains limited to a training scale far smaller than those used by recent supervised depth foundation models, such as the Depth Anything series \cite{yang2024depth,yang2025depth,lin2025depth} and MoGe \cite{wang2025moge}. How SSMDE approaches would perform when scaled to such data regimes remains largely unexplored and is left for future research.

\section{Conclusion}
This study revisited self-supervised depth estimation from the perspective of large-scale, mix-domain training and prediction,
recognizing that the model's tendency to capture dataset-specific structural bias is the key barrier.
MaSS addressed this by coupling a depth encoder with a pretrained VFM and explicitly decoupling the structural representations from the contextual ones to learn scene-generalized structural representations via contrastive learning.
MaSS achieved SoTA accuracy and remarkable out-of-distribution generalization, demonstrating improved scalability with larger and more diverse data.

\begin{ack}
This research was supported in part by the National Key R\&D Program of China under Grant 2025YFE0202900, and by the National Natural Science Foundation of China under Grant 62473288 and Grant 62233013, the Explorers Program of Shanghai (Basic Research Funding) under Grant
26TS1404200, the State Key Laboratory of Human-Machine Hybrid Augmented Intelligence, Xi'an Jiaotong University (No. HMHAI-202406), the Fundamental Research Funds for the Central Universities, NIO University Programme (NIO UP), and the Xiaomi Young Talents Program.

The authors declare no competing financial interests beyond
the funding disclosed above.
\end{ack}

{
    \small
    \bibliographystyle{unsrt}
    \bibliography{egbib}
}

\clearpage
\appendix

\renewcommand{\thefigure}{S\arabic{figure}}
\renewcommand{\thetable}{S\arabic{table}}
\renewcommand{\theequation}{S\arabic{equation}}
\setcounter{table}{0}
\setcounter{figure}{0}
\setcounter{equation}{0}
\renewcommand{\theHequation}{appendix.\arabic{equation}}

\section*{Outline}
In this Supplementary Material, we provide the following:

\begin{itemize}
    \item \textbf{Appendix \ref{sect.add_tech}: Additional Technical Details.}
    Quantitative analyses of the proposed scene-structure discrimination objective, additional training details, and detailed descriptions of the datasets used in this work.

    \item \textbf{Appendix \ref{sect.add_exp}: Supplementary Experimental Results.}
    Supplementary quantitative results under the different alignment protocols, and additional experiments evaluating model generalizability and geometric robustness on in-the-wild, out-of-distribution data, compared with SoTA self-supervised depth estimation methods and geometry foundation models.

    \item \textbf{Appendix \ref{sect.discussions}: Discussion and Future Work.}
    A discussion of the limitations of our method and potential directions for extending it to advance monocular depth estimation and related vision tasks.
\end{itemize}

\section{Additional Technical Details}
\label{sect.add_tech}

\begin{figure*}[h!]
	\centering
	\includegraphics[width=0.99\textwidth]{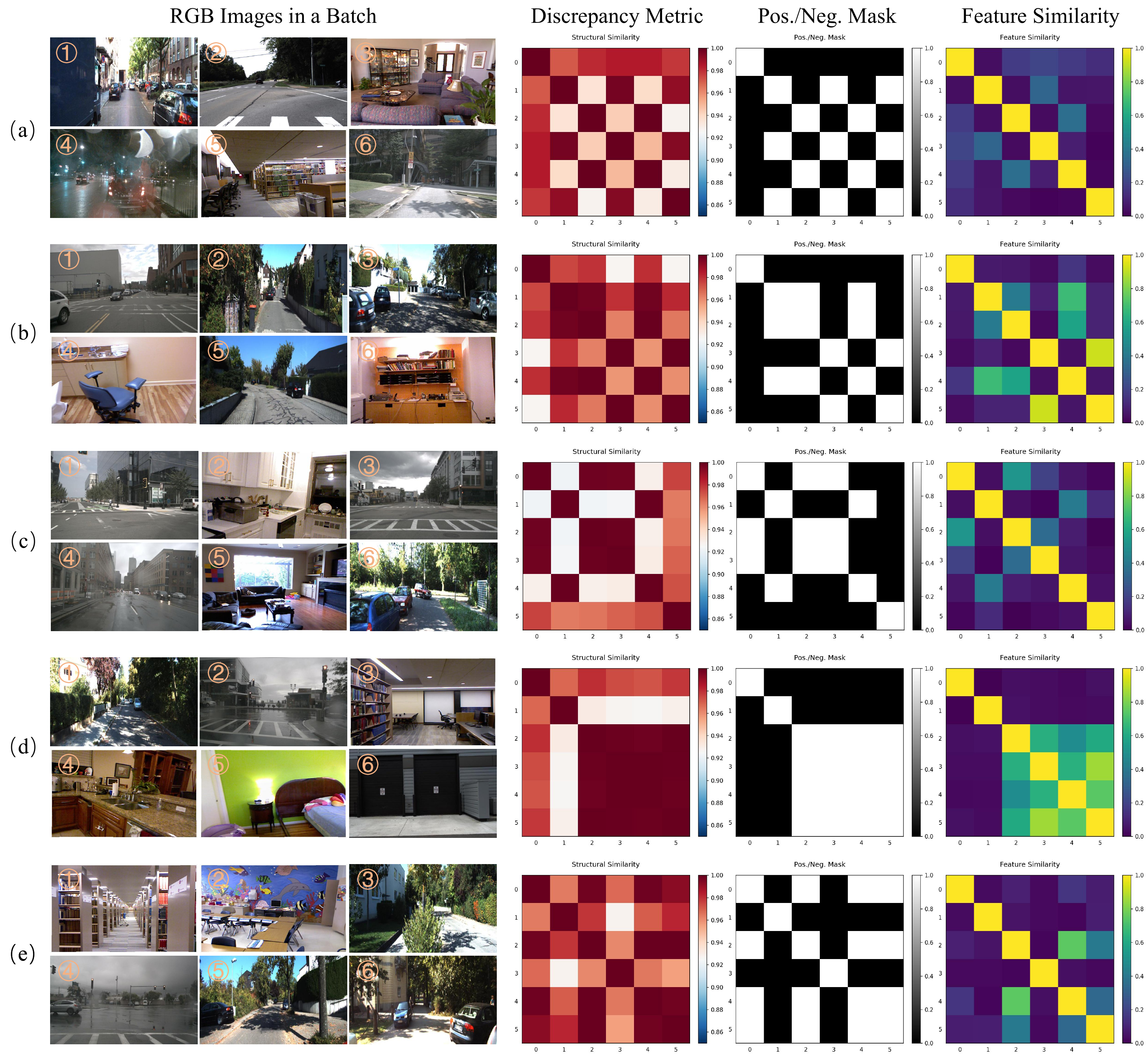}
\caption{Quantitative analyses of the proposed scene-structure discrimination objective for structural representation learning.
}
\label{fig.contrastive}
\end{figure*}

\subsection{Details of Learning to Discriminate Scene Structures}
In the main paper, we define a structural discrepancy metric to determine positive and negative samples within a training batch, thereby formulating a scene-structure discrimination objective for learning structural representations. To illustrate how the metric functions and how the objective guides representation learning, we present several quantitative examples in Fig.~\ref{fig.contrastive}.
The key observations from Fig.~\ref{fig.contrastive} can be summarized as follows:
\\
(1) Our method does not trivially treat each sample as the sole positive instance, nor does it degenerate into discriminating samples based on coarse scene categories (\textit{e.g.}, indoor vs. outdoor).
As shown in Fig. \ref{fig.contrastive}(a), Sample \ding{172} is an outdoor scene but is not considered positive relative to other outdoor samples; this is because its structural layout—a narrow passage flanked by cars and buildings—differs from more open outdoor environments. Fig. \ref{fig.contrastive}(b) provides an inverse case, where a wide-open outdoor scene (Sample \ding{172}) is correctly distinguished from narrow outdoor passages. These observations are further supplemented by Sample \ding{177} in Fig. \ref{fig.contrastive}(c).
Moreover, our method captures shared structural properties across scene categories. For instance, in Fig. \ref{fig.contrastive}(d), the outdoor scene in Sample \ding{177} forms positive pairs with multiple indoor samples due to their similar structure. Similarly, in Fig. \ref{fig.contrastive}(e), the indoor library aisle (Sample \ding{172}) is identified as positive relative to outdoor samples \ding{174}, \ding{176}, and \ding{177}, which share the ``narrow corridor'' structure. Meanwhile, it remains negative to the room scene of Sample \ding{173}. These results demonstrate that our model prioritizes underlying structural similarity over nominal scene labels.
\\
(2) The hyperparameters of the proposed structural discrepancy metric, together with the masking strategy, effectively identify meaningful contrastive relationships. As visualized in the Discrepancy Metric and Pos./Neg. Mask columns, the resulting similarity maps align well with the expected structural relationships described in the above analysis. Meanwhile, the threshold used in our implementation effectively determines the positive and negative sample pairs accordingly.
\\
(3) Finally, within the MaSS framework, the specialized pathway effectively learns structural representations that are discriminative with respect to different scene structures. This is evidenced by the strong consistency between the feature similarities of the learned structural representations and the in-batch structural discrepancy relationships computed from the predicted depth maps.

\subsection{Additional Training Details}

\subsection{Details of Training-Evaluation Datasets}
\textbf{KITTI \cite{geiger2012we}.} Representing standard outdoor driving scenarios, we adopt the widely-used Eigen split \cite{eigen2014depth}, comprising 39,180 monocular image triplets for training, 4,424 for validation, and 697 for testing.
\\
\textbf{DDAD \cite{guizilini20203d}.} To capture dynamic urban scenes with dense traffic, we utilize the DDAD dataset, which is split into a training set of 12,650 images and a test set of 3,950 images.
\\
\textbf{NYUv2 \cite{silberman2012indoor}.} For indoor environments, the NYUv2 dataset is split into 654 testing images for depth evaluation, while the remaining 335 sequences are utilized for training and validation.
\\
\textbf{nuScenes \cite{caesar2020nuscenes}.} Serving as our source for adverse weather and varied lighting conditions, we use 79,760 image triplets from the nuScenes dataset for training, and evaluate the model’s performance on 6,019 front-camera images.

\subsection{Details of Zero-shot Evaluation Datasets}
\textbf{Waymo Open \cite{mei2022waymo}.} Serving as a large-scale outdoor driving benchmark containing diverse urban and suburban scenes, we evaluate models on a test split consisting of 2,216 images, following prior works \cite{sun2023dynamo, zhang2024dcpi}.
\\
\textbf{DrivingStereo \cite{yang2019drivingstereo}.} To evaluate generalization across diverse weather (e.g., rainy and foggy) and lighting conditions in driving scenarios, we use the official test split containing 2,000 images (500 images per weather condition).
\\
\textbf{Cityscapes \cite{cordts2016cityscapes}.} As a benchmark of high-resolution urban street scenes with dense traffic participants, we evaluate on the standard validation set containing 1,525 images, using ground-truth depth from \cite{watson2021temporal}.
\\
\textbf{iBims-1 \cite{koch2020ibims}.} Specifically designed to test indoor depth estimation on complex structures (e.g., transparent objects and fine details), we utilize the official test set of 100 images.
\\
\textbf{SUN-RGBD \cite{song2015sun}.} A large-scale indoor dataset with high scene diversity. We follow the evaluation protocol of \cite{shao2023iebins} using a test split of 5,050 images.
\\
\textbf{7-Scenes \cite{shotton20137scene}.} Collected from a handheld RGB-D camera in various indoor environments, we evaluate on a test split comprising 1,700 images across seven different scenes, following the prior work \cite{bian2021auto}.
\\
\textbf{DIML \cite{cho2021diml}.} To evaluate robustness on purely out-of-distribution data, we use the DIML test split, which contains 500 outdoor images captured in close-range environments whose depth characteristics differ significantly from typical driving datasets.
\\
\textbf{DIODE \cite{vasiljevic2019diode}.} Featuring diverse high-resolution scenes with highly accurate sensor depth measurements, we evaluate on the official test set of 771 images (325 indoor and 446 outdoor). Compared with commonly used indoor benchmarks, the indoor subset spans a larger depth range, while the outdoor subset also mainly consists of closer-range scenes rather than typical driving environments.

\subsection{Data Distribution and Model Generalizability Analyses}

\begin{figure*}[t!]
	\centering
	\includegraphics[width=0.99\textwidth]{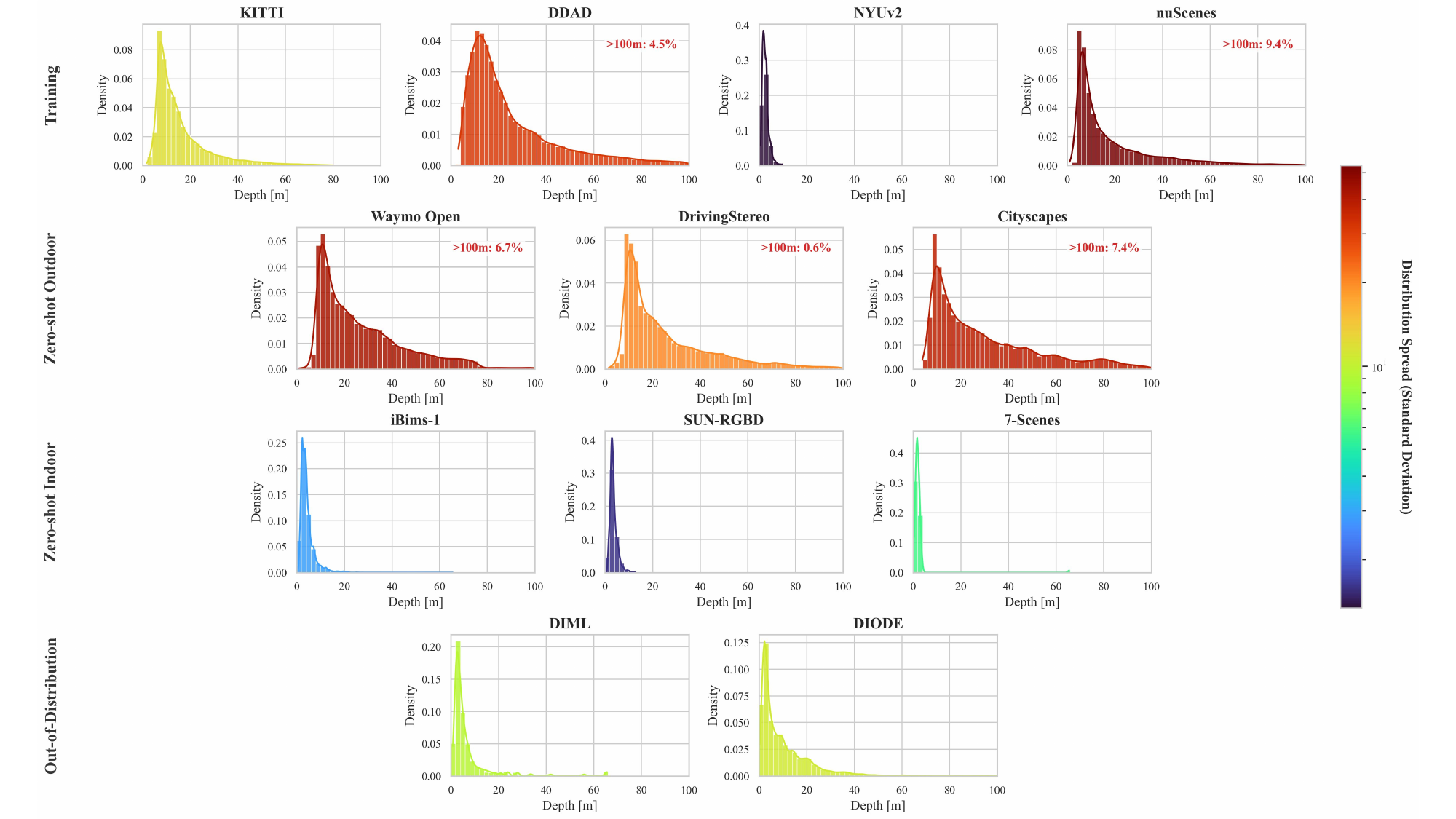}
\caption{Quantitative visualization of the scene structure (depth) distributions across all training and evaluation datasets.
}
\label{fig.depth_dist}
\end{figure*}

\begin{figure*}[t!]
	\centering
	\includegraphics[width=0.99\textwidth]{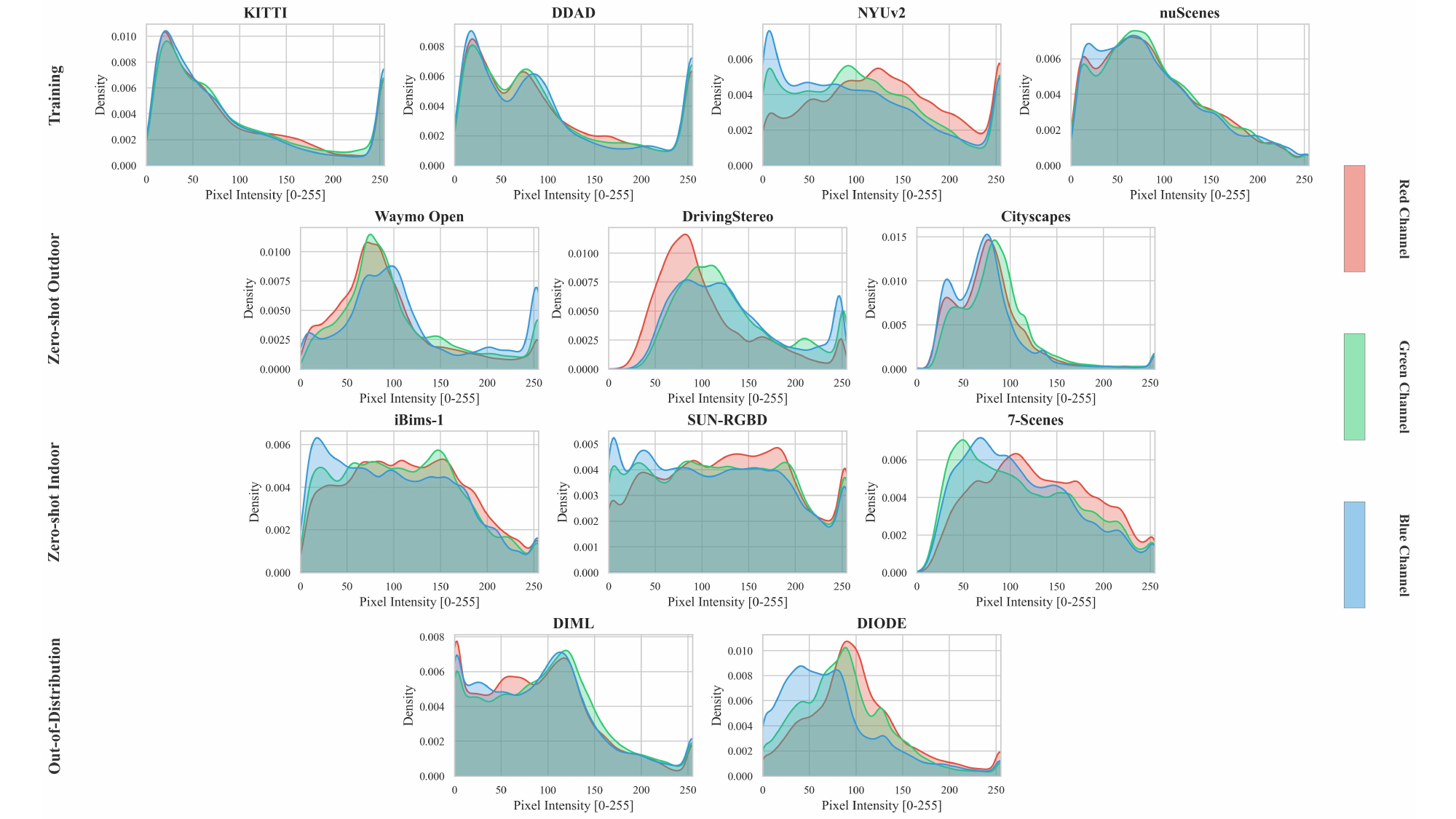}
\caption{Quantitative visualization of the scene appearance (RGB) distributions across all training and evaluation datasets.
}
\label{fig.color_dist}
\end{figure*}

We provide a quantitative analysis of the data distributions across all datasets used in this work to better illustrate how our approach improves the scalability of self-supervised depth learning and enhances model generalizability.
As shown in Fig.~\ref{fig.depth_dist}, the training datasets (KITTI \cite{geiger2012we}, DDAD \cite{guizilini20203d}, and NYUv2 \cite{silberman2012indoor}) exhibit distinct depth distributions. These discrepancies arise from differences between indoor and outdoor environments, as well as variations in field of view (FoV) and sensor configurations during data acquisition~\cite{feng2026panobench}.
In addition, Fig.~\ref{fig.color_dist} presents the RGB distributions of the datasets. Since monocular depth estimation maps RGB inputs to depth outputs, diversity in both domains is crucial for robust generalization. Notably, while nuScenes shares a relatively similar depth distribution with KITTI, its RGB distribution differs significantly.

These observations help interpret the zero-shot evaluation results. For the zero-shot outdoor datasets (Waymo Open \cite{mei2022waymo}, DrivingStereo \cite{yang2019drivingstereo}, and Cityscapes \cite{cordts2016cityscapes}), their depth distributions are similar to those of the driving datasets used for training, whereas their RGB distributions differ noticeably. This suggests that the limited generalization ability of previous methods is not primarily caused by differences in depth ranges. Instead, the performance degradation is largely attributed to appearance shifts, revealing the limited robustness of conventional encoder architectures with restricted pretraining. In contrast, our method learns robust and transferable contextual and structural representations from diverse appearance distributions, substantially improving generalization in depth estimation.
For the zero-shot indoor datasets (iBims-1 \cite{koch2020ibims}, SUN-RGBD \cite{song2015sun}, and 7-Scenes \cite{shotton20137scene}), both the depth and RGB distributions closely align with the NYUv2 training set. This relatively homogeneous distribution favors supervised depth foundation models, as similar scene statistics are already well represented in their large-scale training data. Consequently, these models have already learned similar RGB-to-depth mappings under ground-truth supervision, leading to a larger performance gap between supervised and self-supervised approaches on these datasets.
Finally, DIML \cite{cho2021diml} and DIODE \cite{vasiljevic2019diode} represent clear out-of-distribution (OOD) scenarios. As shown in Figs.~\ref{fig.depth_dist} and \ref{fig.color_dist}, they deviate substantially from the training data in both the input (RGB) and output (depth) domains, making them challenging benchmarks for evaluating the OOD generalization capability of depth estimation methods. Notably, on these datasets, our method achieves performance comparable to that of fully supervised depth foundation models.

\section{Supplementary Experimental Results}
\label{sect.add_exp}

\subsection{Additional Quantitative Results using Different Alignment Protocols}

\begin{table}[t]
\begin{center}
\settablefont
\caption{
Quantitative zero-shot generalization results. Mixed-dataset models ($^{\dagger}$) are evaluated for the image resolutions of 384$\times$640. Supervised foundation models are reported at their pretrained resolutions (518$\times$518 for \cite{yang2025depth, sun2025depth} and 768$\times$768 for \cite{ke2024repurposing, he2025lotus}). Results on the DIML \cite{cho2021diml} dataset are not reported for methods \cite{yang2025depth, sun2025depth} that were pretrained on this dataset. The best results for self-supervised and supervised models are marked in bold type, respectively.}
\label{tb.alignment_compare}
\begin{tabular}{lllcccccccc}
\toprule
\multirow{2}{*}[-0.75ex]{Method} & \multirow{2}{*}{Types}
& \multirow{2}{*}{Align.} & \multicolumn{2}{c}{Waymo} & \multicolumn{2}{c}{DrivingStereo} &
\multicolumn{2}{c}{Cityscapes} & \multicolumn{2}{c}{DIML} \\
\cmidrule(lr){4-5}\cmidrule(lr){6-7}\cmidrule(lr){8-9}\cmidrule(lr){10-11}
& & & Abs~Rel & $\delta_1$ & Abs~Rel & $\delta_1$
& Abs~Rel & $\delta_1$ & Abs~Rel & $\delta_1$  \\
\midrule
DA~V2~\cite{yang2025depth} & SSI* & LSQ*       & 0.084 & 0.912 & 0.103 & 0.908 & 0.088 & 0.920 & -- & --  \\
\midrule
KBR~\cite{spencer2023kick} & SI & Median & 0.156 & 0.806 & 0.139 & 0.825 & 0.111 & 0.864 & 0.105 & 0.902  \\
KBR~\cite{spencer2023kick} & SI & LSQ* & 0.116 & 0.850 & 0.115 & 0.865 & 0.098 & 0.896 & 0.079 & 0.902  \\
\midrule
MaSS$^\dagger$ (Ours)        &SI & Median &  0.123 & 0.867 & 0.100 & 0.905 & 0.103 & 0.896 & 0.104 & 0.890  \\
MaSS$^\dagger$ (Ours)        & SI & LSQ* & 0.088 & 0.908 & 0.088 & 0.916 & 0.087 & 0.917 & 0.089 & 0.922  \\

\midrule
\midrule
\multirow{2}{*}[-0.75ex]{Method} & \multirow{2}{*}{Types}
& \multirow{2}{*}{Align.} & \multicolumn{2}{c}{iBims} & \multicolumn{2}{c}{SUN-RGBD} &
\multicolumn{2}{c}{7-Scenes} & \multicolumn{2}{c}{DIODE} \\
\cmidrule(lr){4-5}\cmidrule(lr){6-7}\cmidrule(lr){8-9}\cmidrule(lr){10-11}
& & & Abs~Rel & $\delta_1$ & Abs~Rel & $\delta_1$
& Abs~Rel & $\delta_1$ & Abs~Rel & $\delta_1$  \\
\midrule
DA~V2~\cite{yang2025depth}    & SSI* & LSQ*     & 0.045 & 0.981 & 0.062 & 0.961 & 0.074 & 0.940 & 0.265 & 0.734  \\
\midrule
KBR~\cite{spencer2023kick} & SI & Median & 0.139 & 0.828 & 0.163 & 0.816 & 0.170 & 0.758 & 0.328 & 0.665  \\
KBR~\cite{spencer2023kick} & SI & LSQ* & 0.104 & 0.890 & 0.121 & 0.867 & 0.117 & 0.865 & 0.255 & 0.679  \\
\midrule
MaSS$^\dagger$ (Ours)        & SI & Median & 0.107 & 0.883 & 0.127 & 0.893 & 0.120 & 0.861 & 0.289 & 0.736  \\
MaSS$^\dagger$ (Ours) & SI & LSQ* & 0.083 & 0.927 & 0.100 & 0.914 & 0.099 & 0.903 & 0.230 & 0.721 \\
\bottomrule
\end{tabular}
\end{center}
\end{table}

The results of the KBR series~\cite{spencer2023kick, spencer2024kick} in Table~\ref{tb.zeroshot} are inconsistent with their originally reported values, as observed on the shared DIODE dataset.
This discrepancy arises because they report results using LSQ alignment in the inverse depth domain (LSQ*), whereas we report results using median alignment in the depth domain.
However, inherent performance gaps exist when using different alignment strategies, as discussed in prior work~\cite{wang2025jasmine}.
In particular, our study further reveals that the choice of alignment domain (depth or inverse depth) can significantly influence performance. To better understand this effect, we analyze the Least-Squares (LSQ) alignment objective in both domains.
Let $\hat{d}_i$ and $d_i$ denote the predicted and reference depth values at pixel $i$, respectively. In the depth domain, LSQ estimates a scale parameter $s$ and a shift parameter $t$ by minimizing
\begin{equation}
(s^{*}, t^{*}) =
\arg\min_{s, t}
\sum_i \left( (s \cdot \hat{d}_i + t) - d_i \right)^2,
\end{equation}
which assigns equal importance to absolute depth errors across all pixels. Since depth uncertainty generally increases with distance, distant regions tend to exhibit larger absolute errors and therefore dominate the optimization process, potentially biasing the global alignment.
In contrast, LSQ$^{*}$ performs alignment in the inverse depth domain. Let $\hat{v}_i = 1/\hat{d}_i$ and $v_i = 1/d_i$ denote the predicted and reference inverse depth values, respectively. The corresponding objective is formulated as
\begin{equation}
(s'^{*}, t'^{*}) =
\arg\min_{s', t'}
\sum_i \left( (s' \cdot \hat{v}_i + t') - v_i \right)^2.
\end{equation}
The inverse depth error can be further expressed as
\begin{equation}
(\hat{v}_i - v_i)^2
=
\frac{(d_i - \hat{d}_i)^2}
{(\hat{d}_i \cdot d_i)^2},
\end{equation}
which shows that LSQ$^{*}$ effectively downweights distant regions through the denominator term. As a result, the influence of high-variance distant pixels is suppressed, preventing the estimated fitting parameters from being dominated by unreliable regions.

As shown in Table~\ref{tb.alignment_compare}, we additionally report results for both KBR~\cite{spencer2023kick} and our method using LSQ* alignment. Results show that this alignment protocol substantially reduces the error metrics. In particular, the DIODE results of KBR under this setting are consistent with those reported in their original paper. Under the same alignment protocol, our method still significantly outperforms prior methods.
In most SSMDE studies, median alignment is a widely adopted and well-justified evaluation protocol. This is because SSMDE inherently learns scale-invariant depth that capture scene geometry without requiring additional shift fitting, whereas LSQ$^{*}$ alignment changes the evaluation objective and no longer faithfully reflects this capability. Therefore, in the main text, we report the results of all self-supervised methods using median alignment.

In addition, the performance gap with DA v2~\cite{yang2025depth} is substantially reduced when also using LSQ*. Specifically, our method outperforms it on several outdoor and out-of-distribution datasets, while only underperforming on relatively easier indoor scenes.
Nevertheless, direct comparisons between models with different invariance properties, such as DA v2 and our method, may not be fully well-defined, even when evaluated under the same LSQ* alignment protocol. This is because DA v2 is explicitly trained to predict scale- and shift-invariant inverse depth, making it inherently compatible with this alignment formulation during both training and inference.  As discussed in their original paper, this design facilitates cross-domain depth learning. In contrast, self-supervised methods do not possess such an inherent advantage.

\subsection{Adapting Depth/Geometry Foundation Models}
As demonstrated in Table~\ref{tb.dfm}, when replacing the original VFM encoder with pretrained depth/geometry foundation models, including DepthAnythingV2 \cite{yang2025depth} and VGGT \cite{wang2025vggt}, MaSS achieves substantial performance gains and also outperforms the original depth prediction heads of these models. This highlights that learning structural representations provides non-trivial benefits even on top of strong depth/geometry-supervised pretraining.
Moreover, since self-supervised methods require only unlabeled monocular videos and inherently predict scale-invariant depth, MaSS can serve as an effective annotation-free adaptation framework for depth foundation models that output affine-invariant depth or disparity, which cannot be directly deployed in many practical scenarios.

\begin{table}[t]
\begin{center}
\settablefont
\caption{Quantitative results on adapting depth/geometry foundation models.
}
\label{tb.dfm}
\begin{tabular}{lllcccccccc}
\toprule
\multirow{2}{*}{Method} & \multirow{2}{*}{Types}
& \multirow{2}{*}{Align.} & \multicolumn{2}{c}{KITTI} & \multicolumn{2}{c}{nuScenes} &
\multicolumn{2}{c}{DrivingStereo} & \multicolumn{2}{c}{Cityscapes} \\
\cmidrule(lr){4-5}\cmidrule(lr){6-7}\cmidrule(lr){8-9}\cmidrule(lr){10-11}
& & & Abs~Rel & $\delta_1$ & Abs~Rel & $\delta_1$
& Abs~Rel & $\delta_1$ & Abs~Rel & $\delta_1$  \\
\midrule
DA V2 \cite{yang2025depth}       & SSI & LSQ & 0.095 & 0.906 & 0.171 & 0.780 & 0.103 & 0.908 & 0.088 & 0.920  \\
DA V2 $+$ Ours & SI & Median & 0.085 & 0.923 & 0.125 & 0.891 & 0.090 & 0.924 & 0.085 & 0.930  \\
\midrule
VGGT \cite{wang2025vggt}       & SI & Median & 0.099 & 0.885 & 0.127 & 0.837 & 0.094 & 0.918 & 0.111 & 0.862  \\
VGGT $+$ Ours & SI & Median & 0.085 & 0.924 & 0.123 & 0.871 & 0.089 & 0.932 & 0.092 & 0.920  \\
\bottomrule
\end{tabular}
\end{center}
\vspace{-0.5em}
\end{table}

\subsection{Generalizability Evaluation on In-the-Wild Data.}

While the previously discussed OOD datasets demonstrate the strong generalization capability of our model, their diversity and domain shift relative to the training data remain limited. To further evaluate robustness, we collect a set of in-the-wild images from online repositories and mobile photography, covering a wide range of real-world challenges.

\begin{figure*}[t!]
	\centering
	\includegraphics[width=0.99\textwidth]{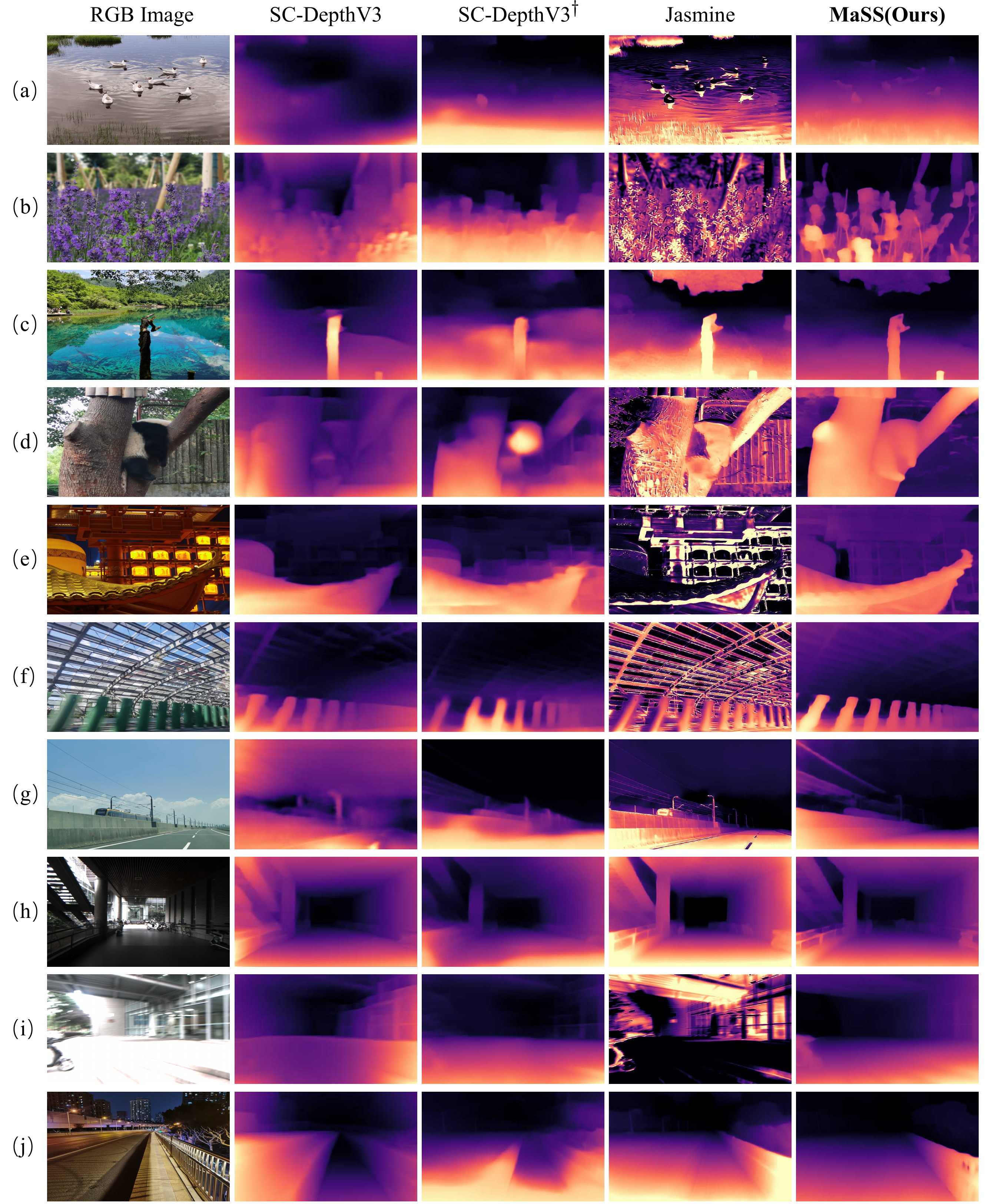}
\caption{Qualitative monocular depth estimation results on in-the-wild, out-of-distribution data.
}
\label{fig.wild_depth}
\vspace{-1em}
\end{figure*}

As illustrated in Fig.~\ref{fig.wild_depth}, the first three rows (a–c) present unstructured natural scenes. Notably, (a) and (c) contain specular reflections on water surfaces, while (b) includes dense vegetation with high-frequency textures.
Rows (d–f) exhibit complex geometries and structures that differ significantly from standard urban or indoor benchmarks, such as the detailed roof tiles in (e) and the repetitive grid structure in (f).
Row (g) includes power lines to evaluate the model’s ability to estimate thin structures.
Rows (h–j) further evaluate model robustness under extreme lighting and motion blur conditions. In particular, row (j) presents a challenging case with strong visual ambiguity at a road junction caused by shadows and lighting effects.

Qualitative comparisons are conducted among SC-DepthV3 \cite{sun2023sc} (trained on NYUv2), SC-DepthV3$^\dagger$, Jasmine \cite{wang2025jasmine} (trained on KITTI), and MaSS$^\dagger$. As shown in Fig.~\ref{fig.wild_depth}, SC-DepthV3 performs unsatisfactorily on complex and unseen structures and tends to interpret scenes with an indoor bias. When trained under the mixed-dataset setting, SC-DepthV3$^\dagger$ further deteriorates, suffering from severe structural confusion caused by the diverse training data. Jasmine represents one of the strongest self-supervised generative depth models. Although it produces detailed predictions and exhibits improved generalization due to diffusion priors, extreme RGB domain shifts can cause the diffusion process to become unstable and produce depth predictions with significant errors. In contrast, our method remains robust across these challenging in-the-wild scenes, producing depth predictions with both fine-grained details and consistent structural layouts.

\begin{figure*}[t!]
	\centering
	\includegraphics[width=0.99\textwidth]{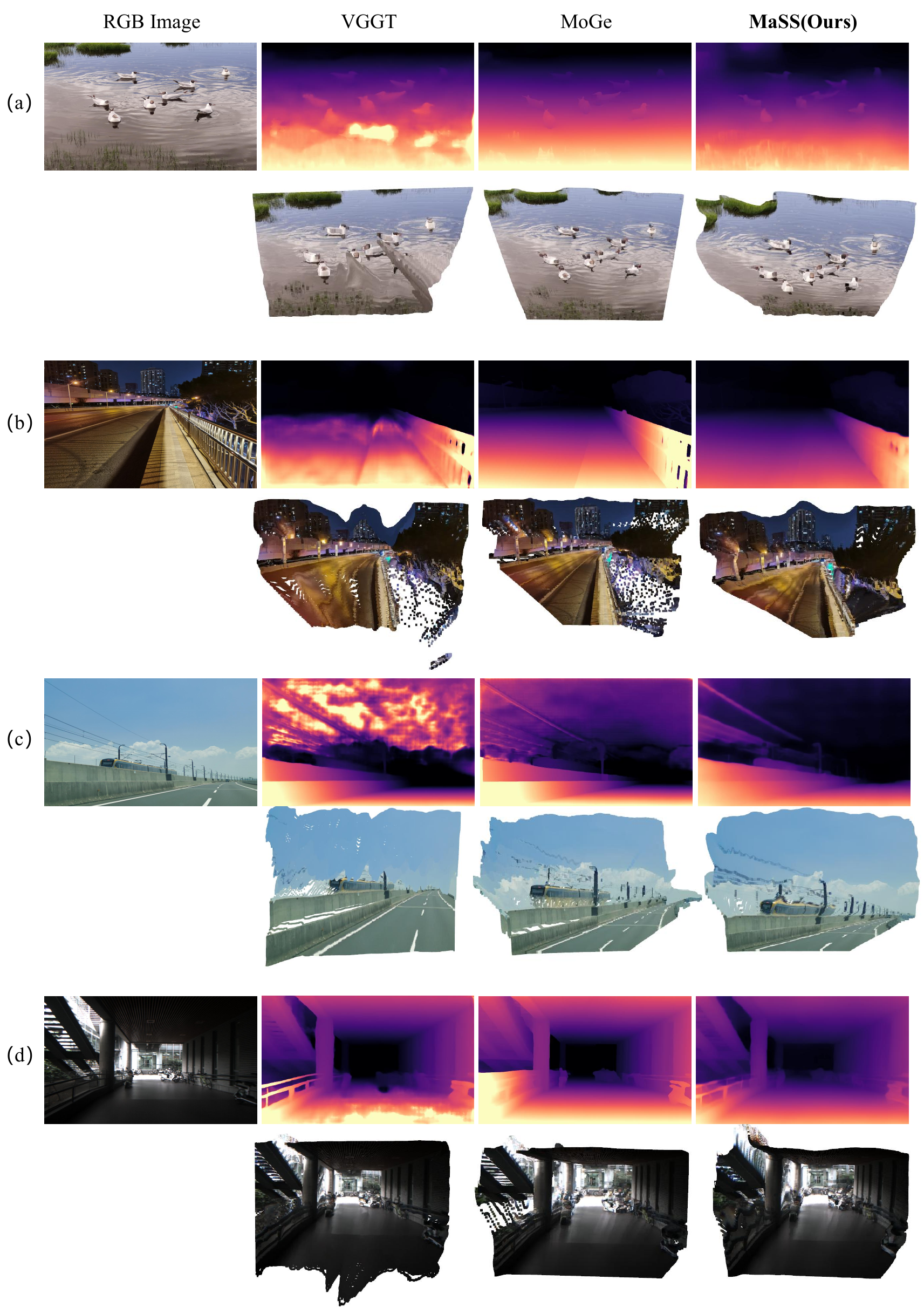}
\caption{Qualitative monocular reconstruction estimation results on in-the-wild, out-of-distribution data.
}
\label{fig.point_cloud}
\end{figure*}

\subsection{Geometric Fidelity Evaluation on In-the-Wild Data.}

Since ground-truth depth is unavailable for these in-the-wild images, quantitative evaluation is infeasible. Instead, we assess the geometric correctness of the predicted depth through monocular 3D reconstruction.

We compare our method with two representative monocular reconstruction approaches, VGGT \cite{wang2025vggt} and MoGe \cite{wang2025moge}. To evaluate geometric fidelity, we synthesize novel viewpoints with slight viewpoint changes relative to the original image. As shown in Fig.~\ref{fig.point_cloud}, VGGT produces relatively noisy depth maps on these in-the-wild scenes, leading to noticeable artifacts in the reconstructed geometry. MoGe represents one of the most robust monocular reconstruction foundation models, benefiting from strong geometric supervision during training, and achieves high-quality reconstructions in most cases, with only minor local inconsistencies observed in (b)-(d). Our method, benefiting from the inherent geometric constraints of self-supervised learning as well as effective scaling with diverse training data, achieves reconstruction results comparable to these prior approaches, with significantly less training data. These results highlight the potential of self-supervised methods for building versatile “DepthAnything”-style models from ubiquitous unlabeled data.

\section{Discussion and Future Work}
\label{sect.discussions}

In addition to increasing the training data scale, as discussed in Sect.~\ref{sect.scale_analyses}, the scalability of SSMDE also depends on the availability of diverse and suitable training data~\cite{wang2026cac}.
Although self-supervision enables learning from imagery collected in the wild, systematically curated datasets covering diverse scenes remain scarce. Existing methods therefore still rely heavily on commonly used datasets originally collected for supervised learning. Constraints imposed by data collection platforms (\textit{e.g.}, vehicles, handheld devices, or UAVs), together with the high cost of sensor calibration and annotation, often limit these datasets to relatively homogeneous scenes, such as typical indoor rooms and common driving environments. This limited diversity may constrain the potential benefits of self-supervision. Such limited diversity may even reduce the potential advantages of self-supervised learning. Systematically curating large and diverse datasets from imagery collected in the wild is therefore a promising direction for advancing SSMDE and extending the benefits of self-supervision to other vision tasks~\cite{zhang2026skytopia, zhou2025gslocalization}.

\end{document}